\documentclass[journal]{IEEEtran}
\usepackage{amsmath,amsfonts}
\usepackage{algorithmic}
\usepackage{array}
\usepackage[caption=false,font=normalsize,labelfont=sf,textfont=sf]{subfig}
\usepackage{textcomp}
\usepackage{stfloats}
\usepackage{url}
\usepackage{hyperref}
\usepackage{verbatim}
\usepackage{graphicx}
\usepackage{cite}
\usepackage{multirow}
\usepackage{makecell}
\usepackage{booktabs}
\usepackage{tabularx}
\usepackage{amssymb}
\usepackage{ulem}
\usepackage{hhline}
\usepackage{titlesec}
\usepackage[table]{xcolor}
\usepackage[ruled,linesnumbered]{algorithm2e}
\def\BibTeX{{\rm B\kern-.05em{\sc i\kern-.025em b}\kern-.08em
    T\kern-.1667em\lower.7ex\hbox{E}\kern-.125emX}}
\usepackage{balance}

\begin{document}
\normalem
\title{A Novel Population Initialization Method via Adaptive Experience Transfer for General-Purpose\\Binary Evolutionary Optimization}
\author{Zhiyuan~Wang,~\IEEEmembership{Student Member,~IEEE},
Shengcai Liu,~\IEEEmembership{Member,~IEEE},
Shaofeng Zhang,
\\\and Ke Tang,~\IEEEmembership{Fellow,~IEEE}
\thanks{Zhiyuan Wang, Shengcai Liu, and Ke Tang are with the Guangdong Provincial Key Laboratory of Brain-inspired Intelligent Computation, Department of Computer Science and Engineering, Southern University of Science and Technology, Shenzhen 518055, China.
(e-mail: wangzy2020@mail.sustech.edu.cn; liusc3@sustech.edu.cn; tangk3@sustech.edu.cn)

Shaofeng Zhang is with Guangdong Provincial Key Laboratory of Brain-inspired Intelligent Computation, Department of Computer Science and Engineering, Southern University of Science and Technology, Shenzhen 518055, China, and Zhongguancun Academy, Beijing 100094, China.
(e-mail: 12445025@mail.sustech.edu.cn)

Corresponding authors: Shengcai Liu and Ke Tang.
}
}


\markboth{IEEE TRANSACTIONS ON EVOLUTIONARY COMPUTATION,~Vol.~XX, No.~X, XX~XXXX}%
{A Novel Population Initialization Method via Adaptive Experience Transfer for General-Purpose Binary Evolutionary Optimization}

\maketitle

\begin{abstract}
    Evolutionary Algorithms (EAs) are widely used general-purpose optimization methods due to their domain independence.
    However, under a limited number of function evaluations~(\#FEs), the performance of EAs is quite sensitive to the quality of the initial population.
    Obtaining a high-quality initial population without problem-specific knowledge remains a significant challenge.
    To address this, this work proposes a general-purpose population initialization method, named mixture-of-experience for population initialization~(MPI), for binary optimization problems where decision variables take values of 0 or 1.
    MPI leverages solving experiences from previously solved problems to generate high-quality initial populations for new problems using only a small number of FEs.
    Its main novelty lies in a general-purpose approach for representing, selecting, and transferring solving experiences without requiring problem-specific knowledge.
    Extensive experiments are conducted across six binary optimization problem classes, comprising three classic classes and three complex classes from real-world applications.
    The experience repository is constructed solely based on instances from the three classic classes, while the performance evaluation is performed across all six classes.
    The results demonstrate that MPI effectively transfers solving experiences to unseen problem classes (i.e., the complex ones) and higher-dimensional problem instances, significantly outperforming existing general-purpose population initialization methods.
\end{abstract}


\section{Introduction}
\label{sec: introduction}

Discrete optimization problems are pervasive in real-world applications like urban planning~\cite{li_tailoring_2024, liu_evolutionary-multimodal_2024, lan_region-focused_2022}, social network analysis~\cite{lu_competition_2015, juarez_recombination_2024}, and industrial manufacturing~\cite{zhang_surrogate-assisted_2021}.
A major subfield of discrete optimization is binary optimization\footnote{Theoretically, any discrete optimization problem can be reformulated as a binary optimization problem.}, where decision variables take values of 0 or 1.
This work focuses on binary optimization due to its fundamental role and widespread presence in real-world applications~\cite{hu_contamination_2010,lu_competition_2015,jiang_smartest_2022,wang2025evolving, dai2023saliencyattack,said2023featureselection}, including adversarial image attack, discrete facility location, and various graph problems (e.g., max cut and max coverage).
Despite its importance, the diversity and complexity across different binary optimization problems make customized algorithm design for each problem challenging, as it relies on experts' domain knowledge and requires substantial human efforts.
Meanwhile, new optimization problems are constantly emerging, including numerous black-box problems that are difficult to model mathematically.
For these problems, such as compiler argument optimization~\cite{jiang_smartest_2022}, obtaining sufficient prior knowledge to design customized algorithms is particularly challenging, highlighting the need for general-purpose methods.

Evolutionary Algorithms (EAs) are general-purpose methods for binary optimization, since they typically treat problems as black boxes, interacting with them only through function evaluations and therefore requiring no problem-specific knowledge.
However, the performance of EAs often diminishes significantly under a limited number of function evaluations~(\#FEs).
This constraint is common in many real-world applications where evaluations are expensive or the time requirement is strict, such as communication network configuration~\cite{li_chance-constrained_2024} and compiler argument optimization~\cite{jiang_smartest_2022}. The constraints of \#FEs limit the EAs' exploration of the solution space, making the quality of the initial population a critical factor for final performance~\cite{segredo_comparison_2018,gong_effects_2023,omidvar_review_2022}.



The ultimate goal of population initialization~(PI) is to enhance the quality of the final solution obtained by EAs.
A core strength of EAs is their generality, i.e., their ability to solve novel problems without relying on problem-specific characteristics~\cite{yu_introduction_2012}.
To preserve this inherent strength, the PI methods themselves should also be general-purpose.
This requirement, however, exposes a significant challenge: existing general-purpose PI methods often fail to generate high-quality initial populations efficiently, creating a performance bottleneck~\cite{tan_evolutionary_2021}.
These methods typically rely on sampling from predefined probability distributions or employing strategic sampling techniques, such as opposition-based initialization~\cite{rahnamayan_opposition-based_2008, keedwell_novel_2018, abdul_halim_algorithm_2024}.
Although widely applicable, their effectiveness is often not satisfactory in scenarios with limited \#FEs. 
Consequently, a critical need exists for an advanced general-purpose PI method that can overcome this limitation and significantly enhance the performance of EAs without relying on problem-specific characteristics. 

A promising strategy to achieve the target above is to leverage knowledge from previously solved problems (source problems) to enhance the performance on the new problems (target problems), a paradigm known as transfer optimization~\cite{feng_towards_2022}.
However, this path is fraught with its own challenges. A primary limitation of many existing transfer optimization methods is their reliance on problem-specific designs and substantial domain expertise to facilitate effective knowledge transfer~\cite{feng_memetic_2015,feng_towards_2022}. This need for specialized design renders these methods largely ineffective for general-purpose scenarios. While a few recent works have achieved greater generality, their applicability is confined to continuous optimization problems~\cite{feng_autoencoding_2017,zhou_learnable_2021}.
To the best of our knowledge, currently there is a significant gap in effective general-purpose PI methods for binary (discrete) optimization, a field encompassing problem classes that are ubiquitous in the real world and represent a primary application domain of EAs.

This work proposes a novel PI method, dubbed \textbf{m}ixture-of-experience for \textbf{p}opulation \textbf{i}nitialization (MPI), for general-purpose binary optimization.
The primary objective of MPI is to enhance the performance of Genetic Algorithms (GAs) by providing a high-quality initial population. 
Specifically, MPI does not rely on problem-specific knowledge, allowing it to be applied directly to a wide range of binary optimization problem classes without adaptation.
Given that GAs stand as classic and highly effective tools for tackling binary optimization problems~\cite{mitchell_introduction_1998}, enhancing their crucial initialization phase in a general-purpose manner represents a valuable contribution to the field.

\begin{figure}[tbp]
	\centering
	\includegraphics[width=0.75\linewidth]{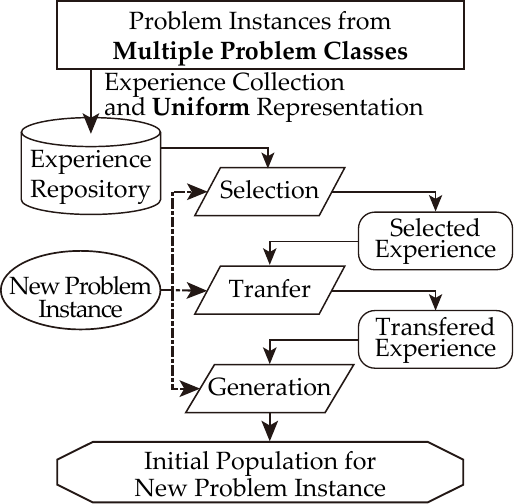}
	\caption{An illustration for the workflow of the proposed MPI method. }
	\label{fig: mpi}
\end{figure}

As illustrated in Fig.~\ref{fig: mpi}, the key novelty of MPI lies in its general-purpose approach for representing, selecting, and transferring optimization experiences during PI for binary optimization.
Specifically, MPI constructs a cross-problem-class experience repository derived from historical solving records and employs a gating network as an experience selection mechanism.
This allows MPI to adaptively leverage solving experiences from diverse problem classes to guide the PI of any new, unseen problem instance.
Consequently, MPI effectively generates high-quality initial populations for the new problem instance, while consuming only a small number of FEs.
This distinguishes MPI from existing PI methods, making it particularly well-suited for \#FE-constrained scenarios and complex cross-problem-class applications.


The main contributions of this work are summarized below.
\begin{itemize}
	\item A novel PI method, namely MPI, is proposed.
    MPI is a general-purpose PI method for solving binary optimization problems with GAs.
    \item Extensive experiments are conducted using two distinct GA variants applied to a diverse set of six binary optimization problem classes.
    These include three classic problem classes and three complex problem classes from real-world applications.
    In the experiments, the experience repository is constructed only using the classic classes, yet the performance evaluation is conducted across all six classes.
    The results demonstrate that MPI successfully transfers solving experiences to unseen complex problem classes as well as higher-dimensional problem instances.
    Furthermore, across all six problem classes, MPI consistently enhances the performance of both GA variants, significantly outperforming existing general-purpose PI methods.
	\item The proposed general-purpose experience representation, selection, and transfer approach not only benefits PI but also provides a new technical route for transfer optimization beyond specific problem classes, potentially inspiring future work in the community.
\end{itemize}

The remainder of this article is organized as follows.
Section~\ref{sec:pop_init} introduces the problem of generating a high-quality initial population for GAs, as well as existing PI methods. 
Section~\ref{sec: offline} and Section~\ref{sec: online} present the proposed MPI method.
Computational studies are presented in Section~\ref{sec:experiments}. Section~\ref{sec:conclusion} concludes the article with discussions. 

\begin{figure*}[htbp]
	\centering
	\includegraphics[width=0.85\textwidth]{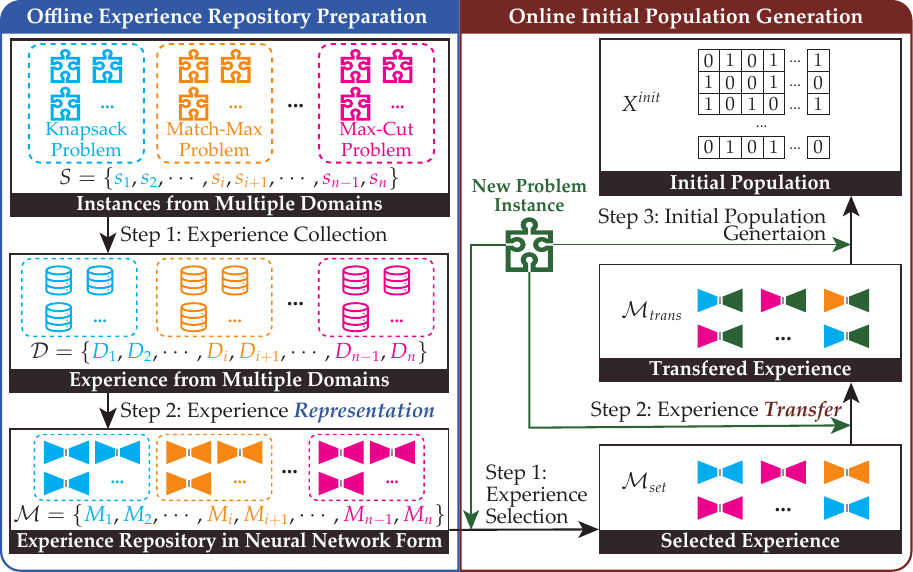}
	\caption{The overall pipeline of MPI,  detailing the steps within the offline and online phases.}
	\label{fig: workflow}
\end{figure*}
\section{Population Initialization for EAs}
\label{sec:pop_init}
\subsection{Notations and Problem Description}

Let \(s\) denote a \(d\)-dimensional optimization problem instance, which is associated with an objective function \(f_s: \{0, 1\}^d\rightarrow\mathbb{R}\). We employ a genetic algorithm (GA), denoted by \(g\), to solve this problem instance.
A PI method, \(h\), is a procedure that generates an initial population \(P_{\text{init}}=h\left(s\right)\). 
The GA, \(g\), then starts from \(P_{\text{init}}\) and evolves, returning the best-found solution, \(\mathbf{x}^*\). Thus, we can represent the entire process as:
\(\mathbf{x}^* = g(h(s), s)\).
The performance of the PI method \(h\) is ultimately evaluated by the quality of this final solution:
\begin{equation}
    \text{Performance}_g\left(h |s\right) = f_s\left(g\left(h\left(s\right), s\right)\right).
\end{equation}
In the optimization process, a budget of \#FEs is allocated jointly to \(h\) and \(g\). In scenarios with limited \#FEs, the key of PI is to generate a high-quality initial population \(P_\text{init}\) with a small number of FEs.

\subsection{Existing Population Initialization Methods}

Existing PI methods can be broadly categorized into two paradigms: random methods and sampling strategy methods~\cite{abdul_halim_algorithm_2024}. Random methods, the most direct approach, create an initial population using pseudo-random number generators~(PRNGs)~\cite{kazimipour_review_2014}, typically drawing from uniform or normal distributions within the search space boundaries. However, due to the lack of guidance toward promising regions, this method is often inefficient at generating high-quality populations, especially when the \#FEs is limited. 

In contrast, sampling strategy methods employ a strategy to guide the sampling process, which can be designed manually or data-driven.
Human-designed sampling strategies, such as opposition-based initialization~\cite{rahnamayan_opposition-based_2008} and Latin hypercube sampling~\cite{iman_latin_2008}, typically aim to produce a population with enhanced spatial uniformity in linear space. The data-driven sampling strategies utilize the information of known solutions to guide the sampling process and generate high-quality solutions, e.g., by constructing surrogate models from a small number of randomly sampled solutions and their corresponding objective values~\cite{keedwell_novel_2018}. Unlike random methods, which operate blindly, both human-designed and data-driven sampling strategies aim to inject prior knowledge or learned information into the population, thereby improving the convergence speed of the search algorithms.

The efficacy of current sampling strategies is notably diminished when operating under strict constraints of \#FEs. Human-designed methods aim to sample more diverse solutions to ensure a thorough exploration of the solution space. This strategy is beneficial when the budget of \#FEs is ample.
However, it can be inefficient under a limited budget as it scatters the exploration process too widely across the solution space. Meanwhile, data-driven methods are similarly inhibited, as the limited budget is insufficient to train and learn an accurate surrogate model. Therefore, both approaches struggle to produce high-quality initial populations for complex binary optimization problems when the budget of \#FEs is limited~\cite{kazimipour_review_2014}. 

\subsection{Experience Transfer for Population Initialization}

The methods discussed above treat each optimization problem instance in an isolated manner, attempting to solve it from scratch. In contrast, transfer optimization offers a more efficient approach by leveraging experiences gained from previously solved problem instances to accelerate the solving process for new ones.
Existing experience transfer mechanisms can be categorized into paradigms of intra-problem-class (within the same problem class)~\cite{feng_memes_2015,zhou_solving_2017,jiang_fast_2021,feng_towards_2022} and cross-problem-class (across different problem classes)~\cite{feng_memetic_2015,feng_autoencoding_2017,zhou_learnable_2021}. Within the context of emphasizing generality, cross-class experience transfer particularly captures our attention. 


Some cross-problem-class transfer methods remain confined to specific, closely related problem classes. Early works in this area required specialized designs, such as experience representation and experience transfer customized to the specific characteristics of routing problems, like the Capacitated Arc Routing Problem (CARP) and the Capacitated Vehicle Routing Problem (CVRP)~\cite{feng_memetic_2015}.
Such reliance on problem-specific knowledge inevitably restricts the method's generality. While some recent studies have proposed general-purpose methods for multi-objective continuous optimization via linear mappings and kernel functions, these methods cannot be applied to binary optimization problems~\cite{feng_autoencoding_2017,zhou_learnable_2021,xue_surrogate-assisted-search_2024}.

To bridge this gap, this work introduces a cross-problem-class PI method for binary problems. We will present the details of our proposed method in the subsequent two sections.

\begin{algorithm}[htbp]
	\small
	\caption{Offline Experience Repository Preparation}
	\label{alg: offline construction}
	\SetKwInput{KwData}{Input}
	\SetKwInput{KwResult}{Output}
	\KwData{Problem instances set \(S\)}
	\KwResult{Experience dataset \(D\), experience repository \(\mathcal{M}\)}
	\SetKw{To}{to}
	\SetKw{Append}{append}
	\SetKw{Into}{into}
	\SetKw{Break}{break}
	\SetKw{Return}{return}
	\SetKwProg{Fn}{Function}{:}{end}
	\SetKwFunction{Sort}{sort}
	\SetKwFunction{Normalize}{normalize}
	\SetKwFunction{Distance}{distance}
	\SetKwFunction{MSE}{MSE}
	\SetKwFunction{MLP}{MLP}
	\SetKwFunction{InsFeature}{ins\_feature}
	\SetKwComment{EmptyLine}{ }{ }
	\SetNoFillComment
	\(\mathcal{D}, \mathcal{M}\leftarrow \left[\right], \left[\right]\)\;
	\For{\(i\leftarrow 1,2,\cdots,n\)}{
        \tcc{Solving Experience Collection}
		\(s_i\leftarrow\) The \(i\)-th instance in \(S\)\;
		\(\mathbf{x}_i^1,\mathbf{x}_i^2,\cdots,\mathbf{x}_i^{m_i}\leftarrow\) Sample \(m_i\) solutions on \(s_i\)\;
		\(y_i^1,y_i^2,\cdots,y_i^{m_i}\leftarrow\) Evaluating \(\mathbf{x}_i^1,\mathbf{x}_i^2,\cdots,\mathbf{x}_i^{m_i}\) on \(s_i\)\;
		\(D_i\leftarrow \left\{\left(\mathbf{x}_i^1,y_i^1\right), \left(\mathbf{x}_i^2,y_i^2\right),\cdots, \left(\mathbf{x}_i^{m_i},y_i^{m_i}\right)\right\}\)\;
        \tcc{Solving Experience Representation}
		\(M_i\leftarrow\) Neural Network with architecture in Fig.~\ref{fig: surrogate}\;
		\(M_i\leftarrow\) Optimize the \(\mathbf{w}^E,\mathbf{w}^D,\mathbf{w}^S\) in \(M_i\) by stochastic gradient descent to minimize the Eq.~(\ref{eq: loss}) with \(D_i\)\;
		Append \(D_i\) into \(\mathcal{D}\)\;
		Append \(M_i\) into \(\mathcal{M}\)\;
	}
    \Return \(\mathcal{D}, \mathcal{M}\)\;
\end{algorithm}
\section{Offline Experience Repository Preparation}
\label{sec: offline}

As illustrated in Fig.~\ref{fig: workflow}, the core idea of MPI is to leverage the solving experiences of previous problem instances (from various problem classes), to facilitate high-quality PI for the GA on any new problem instance.
Specifically, this is achieved through a two-stage process: an offline stage and an online stage.
Crucially, the offline stage is a one-time process.
In this stage, solving experiences from various previously solved problem instances are collected and converted into a uniform representation.
Subsequently, for any incoming new instance in the online stage, MPI selects suitable solving experiences, which are then fine-tuned to adapt to the characteristics of the new instance. A high-quality initial population is ultimately generated for the new instance based on this fine-tuned experience.
A critical distinction in these two stages is the computational budget: the offline stage, being a one-off investment, allows for an ample budget of \#FEs; in contrast, the online stage must operate under a strictly limited budget of \$FEs. This section will focus on detailing the offline stage of the MPI method.


As shown in Fig.~\ref{fig: workflow} and Alg.~\ref{alg: offline construction}, the offline stage consists of two steps: (1) solving experience collection (line~3-6); and (2) solving experience representation (line~7-10). Initially, we assume a set of binary optimization instances \(S=\left\{s_1,s_2,\cdots,s_n\right\}\) from different problem classes and with varying dimensions. 
Crucially, beyond leveraging the inherent pseudo-Boolean solution representation, our approach requires no problem-specific characteristics of the instances in \(S\).

\subsection{Step 1: Solving Experience Collection}
\label{subsec: experience_collection}

In this step, we define the solving experience of a problem instance as a dataset comprising a large number of solution-objective value pairs: \((\mathbf{x}, y)\). 
The most fundamental characteristic of an instance \(s_i\) is its unique solution-to-objective mapping, denoted by the objective function \(f_{s_i}\). While \(f_{s_i}\) often lacks an analytical form, it can be empirically represented by a set of input-output samples. Therefore, our dataset-based definition of experience serves as a direct, static summary of the instance's inherent characteristics, rather than the byproduct of a specific search algorithm.

To acquire this dataset, we sample \(m_i\) solutions for each instance \(s_i \in S\) and evaluate their objective values. For each sampled solution \(\mathbf{x}^i_j\), there is:
\begin{equation}
    y^i_j=f_{s_i}\left(\mathbf{x}^i_j\right),
\end{equation}
where \(f_{s_i}\) is the objective function of instance \(s_i\). This process results in a dedicated dataset \(D_i=\left\{\left(\mathbf{x}^i_j, y^i_j\right)\right\}_{j=1}^{m_i}\) for each instance. The complete experience dataset collection is \(\mathcal{D}=\left\{D_1, D_2, \cdots, D_n\right\}\).



\begin{figure}[htbp]
	\centering
	\includegraphics[width=0.95\linewidth]{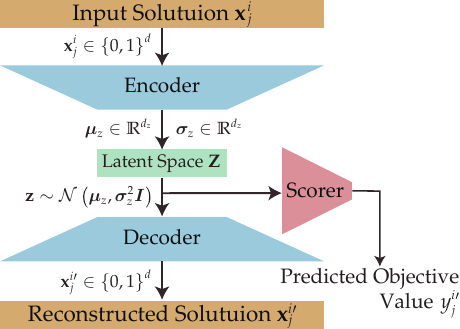}
	\caption{An illustration for the structure of the surrogate model that represents the solving experience for a problem instance.}
	\label{fig: surrogate}
\end{figure}

\subsection{Step 2: Uniform Representation of  Solving Experience}
\label{subsec: experience_representation}


After collecting the experience dataset, we need to convert all the experience datasets in \(\mathcal{D}\) to a uniform representation. In MPI, we employ a variational autoencoder~(VAE)-based neural network as the uniform representation for experiences~\cite{kingma_auto-encoding_2014}. Crucially, this neural network is designed to model only the mapping from solutions to objective values, without incorporating problem-specific constraint information. This design is intentional, as constraints are typically problem-dependent, and encoding them into the model's representation would compromise its generality. To handle constrained optimization problems, we apply constraint handling techniques during data preparation. Any infeasible solutions are transformed into feasible ones, and the objective values of these transformed solutions are used as the training data~(see Section~\ref{subsec: instane_generation}). For the experience dataset \(D_i\) corresponding to instance \(s_i\), we use it to train the neural network, with the weight parameters \(\mathbf{w}_i\) of the resulting model \(M_i\) serving as the parameterized representation of \(D_i\). By fitting all \(D_i\in \mathcal{D}\), we obtain a neural network set \(\mathcal{M}=\left\{M_1, M_2, \cdots, M_n\right\}\), and \(\mathcal{M}\) is the \textbf{experience repository} constructed in the offline stage.

In representing solving experiences, the neural network architecture we employ is illustrated in Fig.~\ref{fig: surrogate}. This VAE-based structure comprises three multilayer perceptron (MLP) components: an encoder \(F_E\) (with weight parameters \(\mathbf{w}_i^E\)), a decoder \(F_D\) (with weight parameters \(\mathbf{w}_i^D\)), and a scorer \(F_S\) (with weight parameters \(\mathbf{w}_i^S\)), while the model's weight parameters \(\mathbf{w}_i\) are the ensemble \(\{\mathbf{w}_i^E, \mathbf{w}_i^D, \mathbf{w}_i^S\}\) of weight parameters from its three components. The model takes solution \(\mathbf{x}^i_j\in \left\{0,1\right\}^d\) as input and outputs reconstructed solution \(\mathbf{x}^{i\prime}_j\in \left\{0,1\right\}^d\) along with predicted objective value \(y^{i\prime}_j\in \mathbb{R}\). Specifically, the encoder \(F_E\) processes \(\mathbf{x}^i_j\) to generate the mean \(\boldsymbol{\mu}_z\in\mathbb{R}^{d_z}\) and standard deviation \(\boldsymbol{\sigma}_z\in\mathbb{R}^{d_z}\) of a \(d_z\)-dimensional multivariate Gaussian distribution \(\mathcal{N}\left(\boldsymbol{\mu}_z,\boldsymbol{\sigma}_z^2\boldsymbol{I}\right)\), which is named latent space \(\mathbf{Z}\). A continuous latent vector \(\mathbf{z}\in\mathbb{R}^{d_z}\) is then sampled from the latent space:  
\begin{equation}  
\label{eq: encoder}  
\begin{aligned}  
    \left[\boldsymbol{\mu}_z,\boldsymbol{\sigma}_z\right]&=F_E\left(\mathbf{x}^i_j, \mathbf{w}_i^E\right)\\  
    \mathbf{z}&\sim\mathcal{N}\left(\boldsymbol{\mu}_z,\boldsymbol{\sigma}_z^2\boldsymbol{I}\right).  
\end{aligned}  
\end{equation}  
The decoder \(F_D\) processes \(\mathbf{z}\) to get the reconstructed solution:  
\begin{equation}  
\label{eq: decoder}  
\begin{aligned}  
    \mathbf{x}^{i\prime}_j&=F_D\left(\mathbf{z}, \mathbf{w}_i^D\right).
\end{aligned}  
\end{equation}  
The scorer \(F_S\) processes \(\mathbf{z}\) to predict the objective value:  
\begin{equation}  
\label{eq: scorer}  
\begin{aligned}  
    y^{i\prime}_j&=F_S\left(\mathbf{z}, \mathbf{w}_i^S\right).
\end{aligned}  
\end{equation}  

When training the VAE-based experience representation \(M_i\) using experience dataset \(D_i\) from instance \(s_i\), the objective function is:  
\begin{equation}  
    \label{eq: loss}  
    \begin{aligned}        
        \min_{\mathbf{w}^E,\mathbf{w}^D,\mathbf{w}^S}\sum_{\left(\mathbf{x}^i_j, y^i_j\right) \in D_i} & \textrm{MSE}\left(\mathbf{x}^i_j, \mathbf{x}^{i\prime}_j\right)+\lambda_1\textrm{MSE}\left(y^i_j, y^{i\prime}_j\right) \\  
        +\lambda_2&\mathbb{D}_\textrm{KL}\left(\mathcal{N}\left(\boldsymbol{\mu}_z,\boldsymbol{\sigma}_z^2\boldsymbol{I}\right),\mathcal{N}\left(\boldsymbol{0},\boldsymbol{I}\right)\right).\\  
    \end{aligned}  
\end{equation}  
Here, \(\mathbf{x}^i_j\) and \(y^i_j\) denote the sampled solution and the corresponding objective value from \(D_i\). Given input \(\mathbf{x}^i_j\) to \(M_i\), \(\mathbf{x}^{i\prime}_j\) and \(y^{i\prime}_j\) represent outputs from the decoder and scorer respectively, while \(\boldsymbol{\mu}_z\) and \(\boldsymbol{\sigma}_z\) are generated by the encoder. MSE denotes mean squared error, and \(\mathbb{D}_\textrm{KL}\) indicates KL-divergence. In Eq.~(\ref{eq: loss}), the first term quantifies reconstruction loss between \(\mathbf{x}^i_j\) and \(\mathbf{x}^{i\prime}_j\), ensuring the encoder captures structural features of input solutions. The second term measures prediction loss between actual objective value \(y^i_j\) and predicted objective value \(y^{i\prime}_j\), ensuring the encoder and scorer adequately fit the objective function of \(s_i\). The third term evaluates KL-divergence between the latent distribution and standard Gaussian, serving as a standard regularization mechanism in VAEs\cite{kingma_auto-encoding_2014} to enforce latent space smoothness.
The primary advantage of using this VAE-based neural network for experience representation lies in its decoupled encoder-decoder architecture and its probability distribution-based latent representation. This design enables efficient experience transfer even with limited data, which will be detailed in Section~\ref{subsec: experience_transfer}.

\section{Online Initial Population Generation}
\label{sec: online}

In the online stage, given a new problem instance, MPI leverages its experience repository to generate a high-quality initial population for the GA to solve it.
As shown in Fig.~\ref{fig: workflow} and Alg.~\ref{alg: online generation}, this stage consists of three main steps: (1) Experience selection, where we select the most suitable subset of experiences from the repository using a gating network (line~1-12); (2) Experience transfer, where we transfer the selected experiences to adapt them to the new problem instance (line~13-23); and (3) Initial population generation, where we leverage the transfer experiences to produce high-quality solutions (line~24-35).
Notably, this entire online stage operates without problem-specific knowledge, treating the new problem instance as a black box.

\subsection{Step 1: Solving Experience Selection}
\label{subsec: experence_selection}

With a limited budget of \#FEs, directly utilizing all experiences in \(\mathcal{M}\), a common practice in previous transfer optimization methods~\cite{zhou_learnable_2021}, could be inefficient. A strategy is therefore required to automatically select an appropriate subset of experiences from \(\mathcal{M}\) for the new instance \(s^{new}\). This concept of automatically selecting multiple suitable experiences is termed mixture-of-experience, which is analogous to the mixture-of-experts paradigm widely used in Large Language Models~\cite{lepikhin_gshard_2021}. To ensure the method's generality, this selection process must treat \(s^{new}\) as a black-box. To address this, MPI computes the relevance between \(s^{new}\) and each experience in \(\mathcal{M}\). These resulting relevance scores serve as input features to a gating network \(G\) (parameterized by \(\mathbf{w}^G\)), which selects the top-\(k\) most suitable experiences.
It should be noted that the gating network \(G\) is an MLP trained during the offline stage, and its training details are provided in Section~\ref{subsec: gating training}.

Given \(s^{new}\), the gating network \(G\) receives a \(3n\)-dimensional input vector, which comprises three relevance coefficients computed between \(s^{new}\) and each of the \(n\) experiences in \(\mathcal{M}\). 
Specifically, relevance coefficients are computed as follows. First, we randomly sample a set of \(e\) solutions \(X^{new}=\left[\mathbf{x}^{new}_1, \mathbf{x}^{new}_2,\cdots,\mathbf{x}^{new}_e\right]\) and calculate their actual objective values \(\mathbf{y}^{new}=\left[y_1^{new}, y_2^{new}, \cdots, y_e^{new}\right]\) on \(s^{new}\).
Next, considering that each experience representation \(M_i \in \mathcal{M}\) serves as a surrogate model for a previously solved problem instance, we input \(X^{new}\) into each of these surrogate models to obtain \(n\) vectors of predicted objective values: \(\mathbf{y}^1, \mathbf{y}^2,\cdots, \mathbf{y}^n\). Here, \(\mathbf{y}^i\) contains the predicted objective values from model \(M_i\) for all solutions in \(X^{new}\). 
To handle dimension mismatches between \(s^{new}\) and the previously solved problem instance, we preprocess \(X^{new}\) by either truncating or zero-padding its dimensions to match those required by each model \(M_i\).
To measure the similarity between the new problem and past experiences, we calculate the Pearson, Spearman, and Kendall's Tau correlation coefficients between the actual objective value vector \(\mathbf{y}^{new}\) and each predicted objective value vector \(\mathbf{y}^i\). This yields three \(n\)-dimensional correlation vectors: \(\mathbf{c}_{\text{Pearson}}\), \(\mathbf{c}_{\text{Spearman}}\), and \(\mathbf{c}_{\text{Kendall}}\). These are then concatenated (denoted by \(\oplus\)) to form the final \(3n\)-dimensional feature vector: \(\mathbf{c}=\mathbf{c}_\text{Pearson}\oplus\mathbf{c}_\text{Spearman}\oplus\mathbf{c}_\text{Kendall}\).
Finally, this feature vector \(\mathbf{c}\) is passed through the gating network \(G\) to produce a score vector \(\mathbf{r}\in\mathbb{R}^n\), where each element represents the suitability of the corresponding experience:
\begin{equation}
    \mathbf{r} = G\left(\mathbf{c}, \mathbf{w}^{G}\right).
\end{equation}
The representations that receive the top-\(k\) scores in \(\mathbf{r}\) are then selected to form the final experience set \(\mathcal{M}_{sel}\).

\begin{algorithm}[htbp]
	\small
	\caption{Online Initial Population Generation}
	\label{alg: online generation}
	\SetKwInput{KwData}{Input}
	\SetKwInput{KwResult}{Output}
	\KwData{New problem instance \(s_{new}\), experience dataset \(D\), experience repository \(\mathcal{M}\), population size \(p\).}
	\KwResult{Generated initial population \(X^{init}\).}
	\SetKw{To}{to}
	\SetKw{Append}{append}
	\SetKw{Into}{into}
	\SetKw{Break}{break}
	\SetKw{Return}{return}
	\SetKwProg{Fn}{Function}{:}{end}
	\SetKwFunction{Sort}{sort}
	\SetKwFunction{Normalize}{normalize}
	\SetKwFunction{Distance}{distance}
	\SetKwFunction{MSE}{MSE}
	\SetKwFunction{MLP}{MLP}
	\SetKwFunction{InsFeature}{ins\_feature}
	\SetKwComment{EmptyLine}{ }{ }
	\SetNoFillComment
        \tcc{\textbf{Step 1:Solving Experience Selection}}
	\(X^{new}\leftarrow\) Sample \(e\) solutions for \(s^{new}\) randomly\;
	\(\mathbf{y}^{new}\leftarrow\) Evaluate \(X^{new}\) on \(s^{new}\)\;
	\For{\(i\leftarrow 1,2,\cdots,n\)}{
		\(M_i\leftarrow\)The \(i\)-th experience representation in \(\mathcal{M}\)\;
		\(X^{\prime}\leftarrow\) Truncate or zero-pad \(X^{new}\) to \(M_i\)'s dimension\;
		\(\mathbf{y}^i\leftarrow\) Evaluate \(X^{\prime}\) on \(M_i\)\;
	}
	\(\mathbf{c}_\text{P},\mathbf{c}_\text{S},\mathbf{c}_\text{K}\leftarrow\) Pearson, Spearman, Kendall's Tau correlation coefficients between \(\mathbf{y}^{new}\) and all \(\mathbf{y}^i\)\;
	\(\mathbf{c}\leftarrow \mathbf{c}_\text{P}\oplus\mathbf{c}_\text{S}\oplus\mathbf{c}_\text{K}\)\tcp*{\(\oplus\) denotes concatenation}
	\(\mathbf{r}\leftarrow G\left(\mathbf{c}, \mathbf{w^{G}}\right)\)\tcp*{\(G\) is a MLP in with weight parameters \(\mathbf{w}^G\) and \(\mathbf{r}\in\mathbb{R}^n\)}
	\(\mathcal{M}_{sel} \leftarrow\) The elements of \(\mathcal{M}\) corresponding to the indices of the top-\(k\) values in \(\mathbf{r}\)\;
	\(\mathcal{M}_{trans}\leftarrow \left[\right]\)\;
        \tcc{\textbf{Step 2:Solving Experience Transfer}}
	\ForEach{\(M\in\mathcal{M}_{sel}\)}{
	\(D\leftarrow\) The corresponding experience data of \(M\) in \(\mathcal{D}\)\;
	\(X, \mathbf{y}\leftarrow\) Sample \(4e\) solutions and corresponding objective values from \(D\) randomly\;
	Sort \(X^{new}\) and \(X\) according to \(\mathbf{y}^{new}\) and \(\mathbf{y}\) respectively\;
	\(e^l\leftarrow\min\left(\textrm{uniqueNum}\left(\mathbf{y}^{new}\right), \textrm{uniqueNum}\left(\mathbf{y}\right)\right)\) \;
	\(X^s_1,X^s_2,\cdots, X^s_{e^l}\leftarrow\) Partition \(X\) into \(e^l\) subsets uniformly, solutions with same fitness in the same \(X^s_i\)\;
	\(X^t_1,X^t_2,\cdots, X^t_{e^l}\leftarrow\) Partition \(X^{new}\) into \(e^l\) subsets uniformly, solutions with same fitness in the same \(X^t_i\)\;
	\(D^f\leftarrow\left(X^s_1\times X^t_1\right)\cup\left(X^s_2\times X^t_2\right)\cup\cdots\cup\left(X^s_{e^l}\times X^t_{e^l}\right)\)\tcp*{\(\times\) denotes Cartesian product}
	\(M^\prime\leftarrow\)Fine-tune \(M\)'s \textbf{Decoder} by stochastic gradient to minimize the Eq.~(\ref{eq: fine-tuning}) with \(D^f\)\;
	Append \(M^\prime\) into \(\mathcal{M}_{trans}\)\;
	}
    \tcc{\textbf{Step 3:Initial Population Generation}}
	\For{\(i\leftarrow 1,2,\cdots,k\)}{
	\(M^\prime_i\leftarrow\) The \(i\)-th experience representation in \(\mathcal{M}_{trans}\)\;
	\(X_{in}\leftarrow\) Sample 10 million solutions for \(M^\prime_i\)\;
	\(X_{out}, \mathbf{y}^\prime\leftarrow M^\prime_i\left(X_{in}\right)\) \;
	Sort \(X_{out}\) according to the \(\mathbf{y}^\prime\)\;
	\(X_i^{gen}\leftarrow\) The top-\(q\) unique solutions in \(X_{out}\)\;
	Evaluate \(X_i^{gen}\leftarrow\) on \(s^{new}\)}\;
	\(X^{init}=X_1^{gen}\cup X_2^{gen}\cup\cdots\cup X_k^{gen}\cup X^{new}\)\;
        \(X^{init}\leftarrow\) Apply an interpolation operator to \(X^{init}\)\;
	\lWhile{\(\left\vert X^{init}\right\vert < p\)}{
		Append a random solution into \(X^{init}\)\hspace{-0.3em}
	}
	\(X^{init}\leftarrow\) The solutions with top-\(p\) objective values in \(X^{init}\)\;
	\Return \(X^{init}\)\;
\end{algorithm}
\subsection{Step 2: Solving Experience Transfer}
\label{subsec: experience_transfer}


After an appropriate subset \(\mathcal{M}_{sel}\) is selected, each experience \(M \in \mathcal{M}_{sel}\) is transferred to the new instance \(s^{new}\). The objective of experience transfer is to establish a mapping between the solution spaces of the source instance \(s\) and target instance \(s^{new}\), enabling high-quality solutions from \(s\) to be reconstructed as high-quality solutions for \(s^{new}\), where \(s\) is the instance corresponding to experience \(M\). With the experience representation detailed in Section~\ref{subsec: experience_representation}, experience transfer in MPI is implemented by fine-tuning the decoder component of the VAE within \(M\) so that when high-quality solutions from \(s\) are provided as input, the fine-tuned decoder outputs corresponding high-quality solutions for \(s^{new}\). The primary challenge of this process is to construct the mapping with minimal \#FEs. To address this, MPI reuses the experience dataset \(D\) in the offline stage and the solution set \(X^{new}\) with corresponding objective values \(\mathbf{y}^{new}\) in the experience selection step as training data. Additionally, a dataset construction method based on solution ranking and Cartesian product operations is employed to enhance the mapping's effectiveness, particularly when training data is limited.

To prepare the fine-tuning dataset for a specific experience representation \(M \in \mathcal{M}_{set}\), we use two sets of solutions. The first set contains the \textit{target solutions} \(X^{new} = \left[ \mathbf{x}^{new}_1, \cdots, \mathbf{x}^{new}_{e} \right]\) and their corresponding objective values \(\mathbf{y}^{new}\), which are generated during the experience selection step (Section~\ref{subsec: experence_selection}). The second set consists of a larger set of \textit{source solutions} \(X = \left[ \mathbf{x}_1, \cdots, \mathbf{x}_{4e} \right]\) with their objective values \(\mathbf{y}\), sampled from \(D\), the experience dataset corresponding to \(M\).
To align solutions of comparable quality, we partition both the source and target solutions into \(e^l\) subsets, where \(e^l\) is the minimum number of unique objective values between the two sets (\(\mathbf{y}\) and \(\mathbf{y}^{new}\)). This partitioning, detailed in Appendix~\ref{append: solution_divides}, groups solutions by objective value as uniformly as possible, ensuring solutions with identical objective value belong to the same subset. The resulting subsets are denoted as \(X^t_1, X^t_2, \cdots, X^t_{e^l}\) for the target and \(X^s_1, X^s_2, \cdots, X^s_{e^l}\) for the source.

The fine-tuning dataset \(D^f\) is constructed by forming the Cartesian product of corresponding source and target solution subsets: \(D^f = \left(X^s_1 \times X^t_1\right) \cup \left(X^s_2 \times X^t_2\right) \cup \cdots \cup \left(X^s_{e^l} \times X^t_{e^l}\right)\). Each pair \(\left(\mathbf{x}_{in}, \mathbf{x}_{out}\right) \in D^f\) consists of a source solution \(\mathbf{x}_{in} \in X\) that serves as the input to model \(M\), and a target solution \(\mathbf{x}_{out} \in X^{new}\) that acts as the ground truth for the decoder's output. A prerequisite for this process is adapting the model architecture. Specifically, the width of the decoder's final layer in \(M\) is adjusted to match the dimensionality of the target solutions \(\mathbf{x}_{out}\), accommodating differences between the source and target problem instances.

The model is then fine-tuned by training the decoder's weight parameters \(\mathbf{w}^{D}\) according to the following objective function:
\begin{equation}
    \label{eq: fine-tuning}
    \min_{\mathbf{w}^D} \sum_{\left(\mathbf{x}_{in}, \mathbf{x}_{out}\right)\in D^f} \mathrm{MSE}\left(\mathbf{x}_{out}, \mathbf{x}_{out}^\prime\right), 
\end{equation}
where \(\mathbf{x}_{out}^\prime\) is the output of the decoder when \(\mathbf{x}_{in}\) is fed into the encoder, and \(\mathrm{MSE}\) denotes the mean square error loss. The resulting fine-tuned model, denoted as \(M^\prime\), encapsulates a mapping from the solution space of instance \(s\) to that of \(s^{new}\). This mapping preserves solution quality so that solutions with high objective value in the source space are transformed into solutioons with high objective value in the target space.

Finally, the set of solving experience representations obtained after transferring each experience representation in \(\mathcal{M}_{set}\) through the above process is denoted as \(\mathcal{M}_{trans}\).

\subsection{Step 3: Initial Population Generation}
\label{subsec: pop_gen}

After experience transfer, we leverage the fine-tuned models in \(\mathcal{M}_{trans}\) to generate a high-quality initial population for \(s^{new}\).

Specifically, for each transferred model \(M_i^\prime \in \mathcal{M}_{trans}\), a large number of solutions (2 million in our experiments) are randomly sampled as input to \(M_i^\prime\), yielding corresponding predicted objective values and reconstructed solutions. The reconstructed solutions are then ranked by their predicted objective values. We select the top-\(q\) unique reconstructed solutions to form a high-quality candidate set \(X^{gen}_i\), where the hyperparameter \(q\) is set to 4 in our experiments. To ascertain their actual performance, all solutions within each generated set \(X^{gen}_i\) are then evaluated on the target instance \(s^{new}\).

Finally, we aggregate the candidate sets from all models along with the solutions from the experience selection step (\(X^{new}\)) to form a comprehensive initial solution set: \(X^{init} = \left(\bigcup_{i=1}^{k} X^{gen}_i\right) \cup X^{new}\).
To further improve the quality of this solution set, an interpolation operator is applied to generate \(q_m\) new solutions based on the existing solutions in \(X^{init}\), which are then added to \(X^{init}\) to augment it.
Details of the interpolation operator are provided in Appendix~\ref{append: interpolation}.
If the size of \(X^{init}\) exceeds the required initial population size, the top-\(p\) solutions, ranked by objective value, are selected to form the initial population, while \(p\) is the population size. Conversely, if \(|X^{init}| < p\), the set is augmented with randomly generated solutions until the population size is met.

\subsection{Gating Network Training}
\label{subsec: gating training}
The gating network \(G\) (parameterized by \(\mathbf{w}^G\)) is trained in a data-driven manner during the offline stage, and this network is then used for any new instance \(s^{new}\). As the training process involves numerous details from the online stage, we defer its description to the end of this section. The guiding principle of this training is to teach \(G\) to select experiences from \(\mathcal{M}\) that are most effective for generating high-quality solutions for any given problem instance.
To achieve this, we use an additional set of problem instances \(S_{G}\) for training. The training process evaluates the quality of a given set of weights \(\mathbf{w}^G\) by iterating through each problem instance \(s \in S_{G}\). For each instance, \(G\) selects a subset of experiences \(\mathcal{M}_{sel}\), which are then used to generate a set of high-quality solutions, denoted as \(X_s^{gen}\). The specific generation procedures are detailed in Section~\ref{subsec: experience_transfer} and Section~\ref{subsec: pop_gen}.
The quality of the generated solutions in \(X_s^{gen}\) provides a supervision signal for training \(\mathbf{w}^G\). Specifically, we optimize \(\mathbf{w}^G\) by maximizing the best objective value of the solutions in \(X_s^{gen}\) across all problem instances in \(S_{G}\). This optimization objective is formally expressed as:
\begin{equation}
    \label{eq: gating_objective}
    \max_{\mathbf{w}^G} \sum_{s\in S_G}\max_{\mathbf{x}\in X^{gen}}f_s\left(\mathbf{x}\right),
\end{equation}
where \(X_s^{gen}\) represents the set of solutions generated for instance \(s\), \(f_s\) is the objective function of instance \(s\). Moreover, since the range of the objective function varies across different instances, it is necessary to normalize \(f_s\left(\mathbf{x}\right)\). For each \(s\in S_G\), we sample 100,000 solutions and evaluate their objective values, taking the maximum \(f_s^{max}\) and minimum \(f_s^{min}\) among them. The normalized \(f_s\left(\mathbf{x}\right)\) is given by:
\begin{equation}
    \label{eq: fitness_normalized}
    f_s\left(\mathbf{x}\right) = \frac{f_s^{ori}\left(\mathbf{x}\right)-f_s^{min}}{f_s^{max}-f_s^{min}}.
\end{equation}
To optimize this black-box objective function in Eq.~(\ref{eq: gating_objective}), we employ PGPE~\cite{sehnke_parameter-exploring_2010}, an Evolution Strategy (ES) method (see Appendix~\ref{append: pgpe} for details). It is worth noting that other black-box continuous optimizers could also be used, as the specific choice of optimizer is not central to our proposed method.

\section{Computational Studies}
\label{sec:experiments}

We conducted a series of experiments to evaluate our proposed method, MPI, focusing on four research questions (\textit{RQs}):
\begin{description}
\item[\textit{RQ1}:] How does MPI perform against state-of-the-art general-purpose PI methods?
\item[\textit{RQ2}:] Can MPI successfully transfer experiences to problem instances with dimensions higher than those in the experience repository?
\item[\textit{RQ3}:] Can MPI successfully transfer experiences to entirely unseen problem classes that do not appear in the experience repository?
\item[\textit{RQ4}:] How do MPI's components, including the gating network, experience transfer, contribute to its effectiveness, and how does the choice of objective function (Eq.~(\ref{eq: gating_objective})) for the gating network training influence the final performance?
\end{description}


Transferring experiences to instances with higher dimensions (\textit{RQ2}) and to instances from unseen problem classes (\textit{RQ3}) is a challenging goal. To evaluate MPI's performance against these challenges, the test set includes instances with higher dimensions than those used to construct the experience repository (for \textit{RQ2}). It also introduces three entirely new problem classes from real-world applications that were unseen during the offline stage (for \textit{RQ3}). The instance sets used across all stages are summarized in Table~\ref{tab: Problem Classes}. Specifically, the experience repository is constructed using only three classic binary problem classes: the One-Max Problem (OM)~\cite{eshelman_crossover_1991}, the Knapsack (KP)~\cite{martello_knapsack_1990}, and the Max-Cut Problem (MC)~\cite{garey_computers_1979}. The test set, in contrast, includes not only new instances of these seen problem classes (OM, KP, MC), but also three complex, unseen classes with black-box objective functions: the Complementary Influence Maximization Problem (CIM)~\cite{lu_competition_2015}, the Compiler Arguments Optimization Problem (CAO)~\cite{jiang_smartest_2022}, and the Contamination Control Problem (CCP)~\cite{hu_contamination_2010, oh_combinatorial_2019}. Further ablation studies, detailed in Section~\ref{subsec: exp_gatenet} and Section~\ref{subsec: exp_obj}, are specifically designed to address \textit{RQ4}.

\begin{table}[htbp]
	\centering
	\caption{Overview of Datasets Used in the Experiments.}
	\resizebox{\linewidth}{!}
	{\begin{tabular}{cccc}
		\toprule[1pt]
        \textbf{Purpose} & \textbf{Problem Classes} & \textbf{Dimensions} & \textbf{\#Instances} \\
		\midrule
		\makecell{\textbf{Experience}\\\textbf{Preparation}\\(Offline)} & \makecell{OM, KP, MC} & 30, 35, 40 & 27 (3 per class/dim)\\
        \midrule
		\makecell{\textbf{Gating Network}\\\textbf{Training}\\(Offline)} & \makecell{OM, KP, MC} & \makecell{40, 60,\\ 80, 100} & 36 (3 per class/dim)\\
        \midrule
		\makecell{\textbf{Evaluation}\\(Online)} & \makecell{OM, KP, MC,\\\textbf{CIM, CAO, CCP}} & \makecell{40, 60,\\ 80, 100} & 72 (3 per class/dim)\\
	\bottomrule[1pt]
	\end{tabular}}
	\label{tab: Problem Classes}
\end{table}


\subsection{Problem Instances and Constraint Handling}
\label{subsec: instane_generation}
Among the six problem classes, KP, MC, CIM, and CCP incorporate constraints.
As noted in Section~\ref{subsec: experience_representation}, MPI does not handle constraints directly; instead, simple constraint-handling techniques are applied.
The details for each problem class, including the objective function, constraint-handling techniques, and instance generation, are provided below.

\subsubsection{One-Max Problem (OM)}

The OM~\cite{eshelman_crossover_1991} used here is a generalized version that minimizes the Hamming distance to a binary reference vector \(\hat{\mathbf{x}}\). The classic OM is a special case where \(\hat{\mathbf{x}}\) is an all-ones vector. For a \(d\)-dimensional instance, the objective is formulated as:
\begin{equation}
    \max f\left(\mathbf{x}\right)=d-D_{\textrm{Hamming}}\left(\hat{\mathbf{x}},\mathbf{x}\right).
\end{equation}
OM has no constraints. An instance is generated by sampling the reference vector \(\hat{\mathbf{x}}\) uniformly at random from \(\{0,1\}^d\)

\subsubsection{Knapsack Problem (KP)}

The KP~\cite{martello_knapsack_1990} is an optimization problem to maximize the total value of selected items subject to a weight capacity \(w_{\text{max}}\). A solution \(\mathbf{x} \in \{0,1\}^d\) indicates the selection of \(d\) items, each with a value \(v_i\) and weight \(w_i\), while \(x_i = 1\) indicating selection of the \(i\)-th item and \(x_i = 0\) otherwise. The problem is formulated as:
\begin{equation}
    \begin{aligned}[t]
        \max f\left(\mathbf{x}\right) &= \sum_{i=1}^d v_i x_i \\
 s.t. \quad w_{\text{max}} &\geq \sum_{i=1}^d w_i x_i 
    \end{aligned}\quad .
\end{equation}
If a solution violates the constraint, it is repaired by iterating through indices to find the point where the cumulative weight exceeds \(w_{\text{max}}\), and setting subsequent \(x_i\) to 0. To generate KP instances, \(\mathbf{v}\) and \(\mathbf{w}\) are randomly sampled from \([0,1]\), sorted to maintain \(\forall i, j \in \{1, 2, \cdots, d\}, v_i > v_j \leftrightarrow w_i > w_j\), and \(w_{\text{max}}\) is set to \(\lambda \sum_{i=1}^{d} w_i\) with \(\lambda\) uniformly sampled at random from \([0.2, 0.8]\).

\subsubsection{Max-Cut Problem (MC)}
The MC~\cite{garey_computers_1979} seeks to partition the vertices of an undirected graph \(\mathcal{G}=(V,E)\) into two sets, \(V_1\) and \(V_2\), to maximize the number of edges between them, where the size of one partition is limited, i.e., \(|V_1| \leq k\) for some \(k \leq d/2\). A solution \(\mathbf{x} \in \{0,1\}^d\) represents this partition, with \(x_i=1\) if vertex \(i \in V_1\) and \(x_i = 0\) if in \(V_2\) . Given the adjacency matrix \(A\), the formulation is:
\begin{equation}
    \begin{aligned}[t]
        \max f\left(\mathbf{x}\right) &= \sum_{i=1}^{d}\sum_{i=1}^{d}x_i\left(1-x_j\right)A_{ij}\\
     s.t. \quad k &\geq \sum_{i=1}^{d}x_i
    \end{aligned}\quad .
\end{equation}
If the constraint is violated, i.e., \(\sum_{i=1}^{d} x_i > k\), a repair procedure retains the first \(k\) nodes assigned to \(V_1\) and reassigns the excess to \(V_2\). Instances are generated by creating \(d\)-node graphs with \(\lambda d^2\) edges, where \(\lambda\) is uniformly sampled from \([0.2, 0.4]\). Any disconnected graphs are regenerated to ensure connectivity. The constraint parameter \(k\) is set to \(\lambda^\prime d\), with \(\lambda^\prime\) independently sampled from \([0.2, 0.4]\).

\subsubsection{Compiler Arguments Optimization Problem (CAO)}

The CAO~\cite{jiang_smartest_2022} derived from software compilation, aims to minimize executable size by selecting an optimal set of compiler arguments. This task is vital for resource-constrained environments, as the arguments' effects are highly source-code-dependent and their complex interactions can be counterproductive, potentially increasing the output size.
 For a \(d\)-dimensional CAO instance, there are \(d\) conditional compiler arguments. The source code is denoted as \(F\), and a solution \(\mathbf{x} \in \{0,1\}^d\) represents the enablement of each argument, with \(x_i = 1\) for enabled and \(x_i = 0\) for disabled. The objective function is defined as:
\begin{equation}
    \label{eq: caop objective}
    \max f\left(\mathbf{x}\right) = -\textrm{CompileSize}\left(F,\mathbf{x}\right).
\end{equation}
The objective value is obtained by compiling the source code \(F\) with the arguments specified by \(\mathbf{x}\). A \(d\)-dimensional instance is generated by randomly selecting a source code \(F\) from cbench or polybench-cpu, and a random subset of \(d\) arguments from the 186 conditional arguments in GCC.

\subsubsection{Complementary Influence Maximization Problem (CIM)}

The CIM~\cite{lu_competition_2015} extends the classical influence maximization problem by incorporating interactions between opinions. In CIM, there is a social network \(\mathcal{G}=\left(V,E,p\right)\), where \(p:E\rightarrow\left[0,1\right]\) assigns influence probabilities to edges, a budget \(k\in\mathbb{Z}^+\), and a pre-defined seed set \(S_A\subseteq V\) for opinion A. The goal is to select a seed set \(S_B\subseteq V\) of size \(k\) for opinion B that maximizes its activation, under interaction dynamics governed by parameters \(q_{A|\emptyset}\), \(q_{A|B}\), \(q_{B|\emptyset}\), and \(q_{B|A}\), as described in~\cite{lu_competition_2015}.
In a \(d\)-dimensional CIM instance, a candidate seed set \(C\subseteq V\) are additionally provided where \(|C| = d\). A solution is represented as a binary vector \(\mathbf{x}\), where \(x_i = 1\) indicates that the \(i\)-th node in \(C\) is selected for \(S_B\), and \(x_i = 0\) otherwise. The formulation is:
\begin{equation}
    \label{eq: cimp objective}
    \begin{aligned}[t]
    \max f\left(x\right)&=\textrm{ActiveNum}\left(\mathcal{G},S_A,C,\mathbf{x}\right) \\
 s.t. \quad k&\geq\sum_{i=1}^{d}x_i
    \end{aligned}\quad .
\end{equation}
The objective is evaluated via Monte Carlo simulation of the influence process detailed in~\cite{lu_competition_2015}. Infeasible solutions are repaired by retaining the first \(k\) selected seeds according to their index order.
An instance is generated using graph \(\mathcal{G}\) randomly selected from the Wiki~\cite{leskovec_predicting_2010}, Facebook~\cite{mcauley_learning_2012}, and Epinions~\cite{leskovec_predicting_2010} datasets. The candidate seed set \(C\) of size \(d\) is chosen randomly from \(V\). The interaction parameters \(\left\{q_{A|\emptyset}, q_{A|B}, q_{B|\emptyset}, q_{B|A}\right\}\) are set to either \(\left\{0.5, 0.75, 0.5, 0.75\right\}\) or \(\left\{0.5, 0.25, 0.5, 0.25\right\}\) randomly, and the maximum seed set size \(k\) is an integer randomly sampled from \(\left[0.2d, 0.6d\right]\).

\subsubsection{Contamination Control Problem (CCP)}

The CCP~\cite{hu_contamination_2010,oh_combinatorial_2019} originates from the necessity to prevent contamination in the food production supply chain. The process involves multiple stages, each capable of introducing contamination. At stage \(i\), applying mitigation measures reduces contamination by a random variable \(\Gamma_i\), at a cost \(c_i\); if no action is taken, the contamination rate remains \(\alpha_i\). A \(d\)-dimensional CCP instance comprises \(d\) stages. A binary solution \(\boldsymbol{x} \in \{0,1\}^{d}\) is defined such that \(x_i = 1\) indicates measures are applied at stage \(i\), and \(x_i = 0\) indicates no measures. The proportion of contaminated food at stage \(i\), denoted \(z_i\), is defined recursively as:\(z_i = \alpha_i (1 - x_i)(1 - z_{i-1}) + (1 - \Gamma_i x_i) z_{i-1}.\)
A key constraint in CCP limits the probability that the contamination exceeds a threshold \(u_i\) at any stage, approximated using \(T\) times Monte Carlo simulations. Following prior work~\cite{oh_combinatorial_2019}, this constraint is incorporated into the objective function via a penalty term \(\frac{1}{T}\sum_{k=1}^{T} 1_{\{z_k > u_i\}}\), leading to the formulation:
\begin{equation}
    \max_{\boldsymbol{x} \in \{0,1\}^{d}} -\left( \sum_{i=1}^{d} \left[c_i x_i + \frac{\rho}{T} \sum_{k=1}^{T} 1_{\{z_k > u_i\}} \right] + \lambda \|\boldsymbol{x}\|_1 \right).
\end{equation}
where \(\lambda\) is a regularization parameter, \(\rho\) is a penalty coefficient and is set to \(\rho=1\) . 
The generation of a \(d\)-dimensional CCP instance follows the settings in ~\cite{oh_combinatorial_2019}. The number of Monte Carlo simulations is set to \(T = 100\). The contamination limitation is set to \(u_i = 0.1\). The \(\lambda\) is chosen from \(\{0, 10^{-2}\}\). The random variables are distributed as \(\alpha \sim \mathrm{Be}(1, 17/3)\), \(\Gamma \sim \mathrm{Be}(1, 7/3)\), and the initial contamination follows \(z_0 \sim \mathrm{Be}(1, 30)\), while \(\mathrm{Be}\) means beta distributions.


\subsection{Experimental Protocol and Compared Methods}
\label{subsec: exp_protocol}
To answer our \textit{RQs}, we evaluate MPI against representative PI methods using two distinct GA optimizers. The first optimizer is a conventional GA with elitism, referred to as GA-Elite, which employs single-point crossover and random flip mutation. The second optimizer is Biased Random-Key Genetic Algorithm (BRKGA)\cite{londe_biased_2025}. 
For all experiments, the population size is set to 20, which provides a balance between population diversity and the number of generations under the 800 FEs budget. GA-Elite retains 1 elite individual, generates 20 crossover offspring per generation, and applies a mutation rate of 0.001. BRKGA is configured with 4 elite individuals, 14 crossover and 2 mutant offspring, and an elite bias of 0.7, following the default component ratios recommended in its open-source implementation~\cite{blank_pymoo_2020}.

The performance is evaluated under the budget of 800 FEs, which is chosen as an intermediate value reflecting the 300-1000 range commonly adopted in previous similar works\cite{song_kriging-assisted_2021,hao-evaluated_2025,gonzalez-duque_survey_2024,wei_distributed_2023}, to simulate scenarios with limited \#FEs.
To ensure a fair comparison, any FEs consumed during PI are subtracted from the total budget, leaving the remainder for the subsequent optimization phase.
For statistical robustness, each experiment is repeated 30 times with different random seeds, and all comparisons are conducted using the Wilcoxon Rank-Sum test. Given that the test set comprises 72 instances, presenting per-instance results in the main paper is impractical. Consequently, for the primary analysis at 800 FEs, we report the statistical hypothesis test results aggregated by problem class in the main text and provide the detailed per-instance results in the supplementary material. Furthermore, to examine performance variation trends, we extend the analysis to a budget of 1600 FEs and report the statistical comparison results aggregated across all instances.

We compare MPI against four state-of-the-art or standard PI methods. The parameter settings for each are detailed below. 
\begin{itemize}
    \item \textbf{MPI (Ours)}: MPI generates \(e=64\) initial solutions and selects \(k=12\) experiences. It generates \(q=4\) solutions for each fine-tuned experience. Finally, \(20\) solutions are produced via interpolation. This process consumes a total of \(64 + 12 \times 4 + 20 = 132\) FEs for PI. The top-\(20\) solutions form the initial population.
    \item \textbf{Rand}: A baseline that samples \(20\) solutions from a uniform binomial distribution. It consumes \(20\) FEs.
    \item \textbf{OBL}\cite{rahnamayan_opposition-based_2008}: Opposition-based learning strategy. It samples \(10\) random solutions and then generates another \(10\) by taking their bit-wise complements. It consumes \(20\) FEs.
    \item \textbf{SVM-SS}\cite{keedwell_novel_2018}: A surrogate-assisted sampling strategy. To match MPI's budget, 132 FEs were allocated for PI. It begins by randomly sampling and evaluating \(20\) solutions and training an SVM model. Then, it iteratively identifies a promising solution and trains the SVM model to guide the search until the budget is exhausted. The final \(20\) best solutions are selected.
    \item \textbf{KAES}\cite{zhou_learnable_2021}: A transfer optimization method for continous optimiztion, which we adapted for binary problems by binaryizing the continuous solutions. It is also allocated 132 FEs. It samples 105 initial solutions and then generates 27 more by transferring knowledge from the experience dataset \(\mathcal{D}\), finally selecting the top-\(20\).
\end{itemize}

Experiments were conducted on a server cluster with \texttt{AMD EPYC} CPUs and \texttt{Nvidia A6000/A30} GPUs, running Ubuntu 22.04. The source code and benchmark instances for each problem class have been anonymously open-sourced at \url{https://github.com/AnonymousSubmitBot/EMPI}.

\begin{table*}[htbp]
  \centering
  \caption{Performance Comparison between MPI and Baselines by Problem Class. \#W-D-L: Win-draw-loss Counts from Wilcoxon Rank-Sum Tests (\(p=0.05\); W: MPI Significantly Better, D: No Significant Difference, L: the Baseline Significantly Better). \#Avg.\(\uparrow\): the Number of Instances where MPI has Higher Average Performance. Problem Classes with the Baseline Performs Better on More Instances are Highlighted in \underline{italic and Underline}.}
    \begin{tabular}{ccccccccccc}
    \toprule[1pt]
    \multirow{2}[1]{*}{\textbf{GA}} & \multirow{2}[1]{*}{\makecell{\textbf{Seen}\\\textbf{Class}}} & \multirow{2}[1]{*}{\makecell{\textbf{Problem}\\\textbf{Class}}} & \multicolumn{2}{c}{\textbf{MPI VS. Rand}} & \multicolumn{2}{c}{\textbf{MPI VS. SVM-SS}} & \multicolumn{2}{c}{\textbf{MPI VS. OBL}} & \multicolumn{2}{c}{\textbf{MPI VS. KAES}} \\
    \cmidrule(lr){4-5}\cmidrule(lr){6-7}\cmidrule(lr){8-9}\cmidrule(lr){10-11}
    & & & \textbf{\#W-D-L} & \textbf{\#Avg.}\(\uparrow\) & \textbf{\#W-D-L} & \textbf{\#Avg.}\(\uparrow\) & \textbf{\#W-D-L} & \textbf{\#Avg.}\(\uparrow\) & \textbf{\#W-D-L} & \textbf{\#Avg.}\(\uparrow\) \\
    \hhline{-----------}
      \multirow{7}[3]{*}{\textbf{GA-Elite}} 
      & \multirow{3}[2]{*}{\textbf{True}}
      & MM & 12-0-0 & 12 & 12-0-0 & 12 & 12-0-0 & 12 & 7-3-2 & 8 \\
      & & MC & 12-0-0 & 12 & 5-5-2 & 7 & 12-0-0 & 12 & 8-2-2 & 10 \\
      & & KP & 12-0-0 & 12 & 5-6-1 & 9 & 9-3-0 & 11 & 7-4-1 & 10 \\
      \hhline{~----------} 
      & \multirow{3}[2]{*}{\textbf{False}}
      & CCP & 12-0-0 & 12 & 12-0-0 & 12 & 12-0-0 & 12 & 9-3-0 & 11 \\
      & & CIM & 12-0-0 & 12 & 7-3-2 & 7 & 11-1-0 & 12 & 9-3-0 & 11 \\
      & & CAO & 9-2-1 & 11 & 3-8-1 & 8 & 9-1-2 & 10 & 3-7-2 & 8 \\
      \hhline{~----------} 
      & \multicolumn{2}{c}{\textbf{Total}} & 69-2-1 & 71 & 44-22-6 & 55 & 65-5-2 & 69 & 43-22-7 & 58 \\
      \hhline{>{\global\arrayrulewidth=0.8pt}----------->{\global\arrayrulewidth=0.4pt}}
      \multirow{7}[3]{*}{\textbf{BRKGA}}
      & \multirow{3}[2]{*}{\textbf{True}} 
      & MM & 6-6-0 & 12 & 4-8-0 & 10 & 6-6-0 & 10 & 4-7-1 & 8 \\
      & & MC & 4-7-1 & 6 & \textit{\underline{2-4-6}} & \textit{\underline{4}} & 3-6-3 & 7 & 5-4-3 & 6 \\
      & & KP & 6-6-0 & 9 & 6-5-1 & 9 & 5-7-0 & 10 & 4-6-2 & 7 \\
      \hhline{~----------} 
      & \multirow{3}[2]{*}{\textbf{False}}
      & CCP & 7-5-0 & 11 & 8-4-0 & 10 & 8-3-1 & 9 & 3-8-1 & 8 \\
      & & CIM & 7-4-1 & 10 & 5-4-3 & 7 & 5-7-0 & 8 & 3-8-1 & 8 \\
      & & CAO & 4-6-2 & 9 & 5-5-2 & 8 & 4-6-2 & 8 & 4-5-3 & 8 \\
      \hhline{~----------}  
      & \multicolumn{2}{c}{\textbf{Total}} & 34-34-4 & 57 & 30-30-12 & 48 & 31-35-6 & 52 & 23-38-11 & 45 \\
    \hhline{>{\global\arrayrulewidth=1.0pt}----------->{\global\arrayrulewidth=0.4pt}}
    \end{tabular}
  \label{tab: VS MPI Each Class}
\end{table*}

\begin{figure*}[htbp]
    \centering
    \subfloat[]{\includegraphics[width=0.242\textwidth]{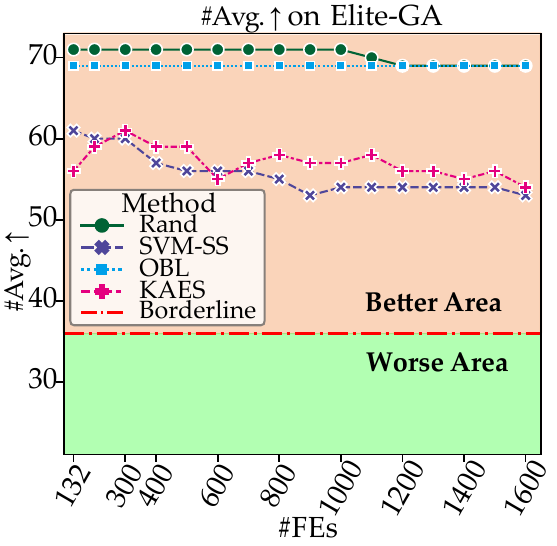}
    \label{fig: mpi elitega avg}}
    \subfloat[]{\includegraphics[width=0.242\textwidth]{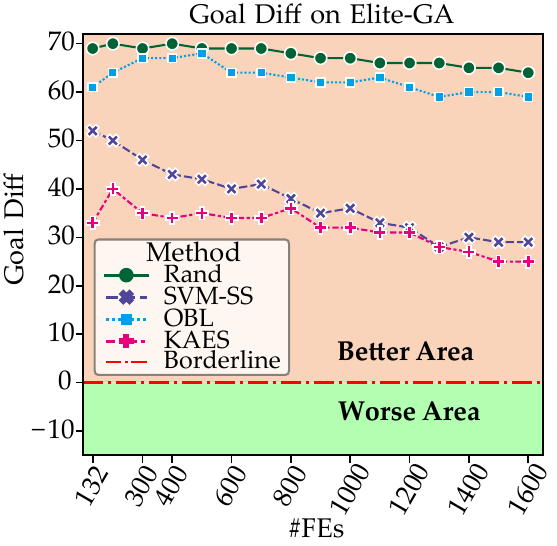}
    \label{fig: mpi elitega goal diff}}
    \subfloat[]{\includegraphics[width=0.242\textwidth]{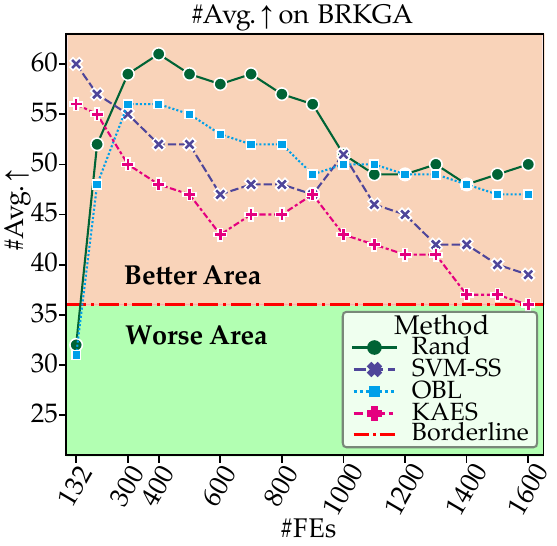}
    \label{fig: mpi brkga avg}}
    \subfloat[]{\includegraphics[width=0.242\textwidth]{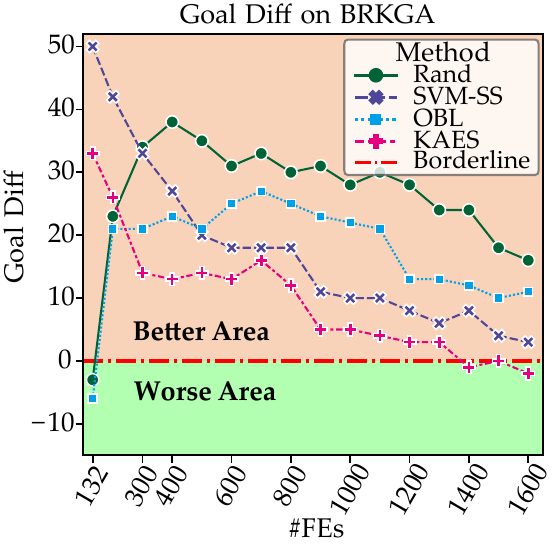}
    \label{fig: mpi brkga goal diff}}
    \caption{Performance comparison of \textbf{MPI VS. the Baselines} across \#FE budgets. X-axis: \#FEs; Y-axis: performance metrics. \#Avg. \(\uparrow\): the number of instances where the MPI's average objective value of the best solution (from 30 runs) exceeds the Baseline's. Goal Diff: Win number minus Loss number from statistical testing (W-D-L results) on best solutions over 30 runs. Each subplot includes a red horizontal line dividing the area into Better (MPI outperforms the Baseline) and Worse (the Baseline outperforms MPI) areas.
    The subfigures present: (a) \#Avg. \(\uparrow\) metric on Elite-GA; (b) Goal Diff metric on Elite-GA; (c) \#Avg. \(\uparrow\) metric on BRKGA; (d) Goal Diff metric on BRKGA.}
    \label{fig: mpi steps}
\end{figure*}

\subsection{Test Results and Analysis}

This section presents the experimental results, comparing the proposed MPI method against state-of-the-art baselines to address \textit{RQ1} (Effectiveness), \textit{RQ2} (Experience transfer across dimensions), and \textit{RQ3} (Experience transfer across problem classes). The primary results are presented in Table~\ref{tab: VS MPI Each Class} and Table~\ref{tab: VS MPI Each Dim}, which detail the statistical comparisons against baselines aggregated by problem class and dimension, respectively. Additionally, Fig.~\ref{fig: mpi steps} illustrates the performance trend of all methods on all test instances as the budget of \#FEs is extended to 1600.


\subsubsection{Results for Effectiveness}
The evaluation of MPI's effectiveness (\textit{RQ1}) begins with a comparison against MPI and baseline methods. The results, summarized in Table~\ref{tab: VS MPI Each Class} (with detailed instance-level results available in the supplementary material), demonstrate MPI's strong performance. When integrated with GA-Elite, MPI consistently outperforms all baselines on all problem classes. Specifically, compared to Rand, MPI achieves a higher mean performance on 71 out of 72 instances, with 69 of those being statistically significant. Against SVM-SS and KAES, MPI demonstrates a higher mean performance on 55 out of 58 instances, respectively, with 44 and 43 of those comparisons showing statistical significance, respecitvely.
When integrated with BRKGA, an optimizer with inherently stronger performance, MPI maintains a performance advantage over the baselines, although this advantage is slightly less pronounced compared to the GA-Elite results. While it is observed that SVM-SS achieves the best performance on the Max-Cut problem, MPI still significantly outperforms the other baselines on this problem class. Therefore, this specific result should be attributed to the exceptional performance of SVM-SS on Max-Cut, rather than a general weakness of MPI for this problem class.

Fig.~\ref{fig: mpi steps} further reveals the performance dynamics as the \#FEs budget increases, confirming that MPI's dominance is maintained across an extended evaluation budget.
After its initial setup phase (132 FEs), the figure shows that MPI consistently maintains a performance advantage over all other methods.
Although MPI's advantage may gradually narrow as the \#FEs budget increases, it achieves substantial performance improvements within the 300-1000 FEs budget range, which is common to many \#FE-constrained scenarios~\cite{song_kriging-assisted_2021,hao-evaluated_2025,gonzalez-duque_survey_2024,wei_distributed_2023}. 
This narrowing trend is reasonable, as a sufficient \#FEs budget allows EAs to explore the solution space more thoroughly, naturally reducing the impact of the initial population on the final solution quality.
Consequently, MPI's primary advantage lies in scenarios where the \#FEs budget is limited.
These findings collectively confirm MPI's superior performance as a general-purpose PI method. MPI's superior performance can be attributed to its adaptive design for \#FEs-constrained scenarios, including the gating network for experience selection and the VAE-based experience transfer method.

\begin{table}[htbp]
  \centering
  \caption{Performance Comparison between MPI and Baselines by Problem Dimensions. \#W-D-L: Win-draw-loss Counts from Wilcoxon Rank-Sum Tests (\(p=0.05\); W: MPI Significantly Better, D: No Significant Difference, L: the Baseline Significantly Better). \#Avg.\(\uparrow\): the Number of Instances where MPI has Higher Average Performance. Dimensions with the Baseline Performs Better on More Instances are Highlighted in \underline{italic and Underline}.}
    \resizebox{\linewidth}{!}{\begin{tabular}{@{}c@{}c@{}c@{}c@{}c@{}c@{}c@{}c@{}c@{}c@{}}
    \toprule[1pt]
    \multirow{2}[1]{*}{\textbf{GA}} & \multirow{2}[1]{*}{\textbf{Dim}} & \multicolumn{2}{c}{\textbf{MPI VS. Rand}} & \multicolumn{2}{c}{\textbf{MPI VS. SVM-SS}} & \multicolumn{2}{c}{\textbf{MPI VS. OBL}} & \multicolumn{2}{c}{\textbf{MPI VS. KAES}} \\
    \cmidrule(lr){3-4}\cmidrule(lr){5-6}\cmidrule(lr){7-8}\cmidrule(lr){9-10}
    & & \textbf{\#W-D-L} & \textbf{\#Avg.}\(\uparrow\) & \textbf{\#W-D-L} & \textbf{\#Avg.}\(\uparrow\) & \textbf{\#W-D-L} & \textbf{\#Avg.}\(\uparrow\) & \textbf{\#W-D-L} & \textbf{\#Avg.}\(\uparrow\) \\
    \hhline{----------}
      \multirow{5}[1]{*}{\textbf{GA-Elite}} 
          & 40    & 16-1-1 & 17    & 11-4-3 & 13    & 16-0-2 & 16    & 7-6-5 & 12 \\
          & 60    & 17-1-0 & 18    & 10-6-2 & 13    & 16-2-0 & 18    & 12-5-1 & 15 \\
          & 80    & 18-0-0 & 18    & 9-8-1 & 12    & 17-1-0 & 17    & 10-7-1 & 14 \\
          & 100   & 18-0-0 & 18    & 14-4-0 & 17    & 16-2-0 & 18    & 14-4-0 & 17 \\
      \hhline{~---------} 
      & \textbf{Total} & 69-2-1 & 71 & 44-22-6 & 55 & 65-5-2 & 69 & 43-22-7 & 58 \\
      \hhline{>{\global\arrayrulewidth=0.8pt}---------->{\global\arrayrulewidth=0.4pt}}
      \multirow{5}[1]{*}{\textbf{BRKGA}}
          & 40    & 10-6-2 & 13    & 10-3-5 & 11    & 8-7-3 & 13    & 5-10-3 & 9 \\
          & 60    & 6-12-0 & 13    & 4-12-2 & 11    & 7-10-1 & 12    & 6-10-2 & 13 \\
          & 80    & 5-11-2 & 13    & 6-8-4 & 11    & 8-9-1 & 11    & \textit{\underline{5-7-6}} & \textit{\underline{8}} \\
          & 100   & 13-5-0 & 18    & 10-7-1 & 15    & 8-9-1 & 15    & 7-11-0 & 15 \\
      \hhline{~---------}  
      & \textbf{Total} & 34-34-4 & 57 & 30-30-12 & 48 & 31-35-6 & 52 & 23-38-11 & 45 \\
    \hhline{>{\global\arrayrulewidth=1.0pt}---------->{\global\arrayrulewidth=0.4pt}}
    \end{tabular}}
  \label{tab: VS MPI Each Dim}
\end{table}

\subsubsection{Transfer Experience to Instances with Higher Dimensions}

To further explore MPI's ability to transfer experience to instances with higher dimensions (RQ2), Table~\ref{tab: VS MPI Each Dim} presents the statistical comparison results aggregated by instance dimension. The results show that MPI demonstrates a significant advantage over the baseline methods across all test dimensions (40, 60, 80, and 100), with one exception: at dimension 80, KAES outperformed MPI on more instances. Although the experience repository included experiences from 40-dimensional instances, MPI did not exhibit a more pronounced performance advantage on the 40-dimensional test instances compared to those at higher, unseen dimensions. In fact, in some comparisons against the baselines, MPI's advantage was greater at higher dimensions than at dimension 40, especially at dimension 100. The reason is possibly that as the problem dimension increases, the search space expands, and the quality of the initial population exerts a more significant impact on the final solution. These findings confirm that MPI effectively transfer experience to novel problem instances with dimensions significantly higher than those in the experience repository.

\subsubsection{Transfer Experience to Instances from Unseen Problem Class}

Table~\ref{tab: VS MPI Each Class} also provides strong evidence that MPI can effectively transfer experience to instances from unseen problem classes (RQ3). MPI demonstrates a significant performance advantage on the instances of all three unseen problem classes (CIM, CAO, and CCP). Although its advantage on CAO instances is slightly smaller than on the other five problem classes when using GA-Elite, this is not due to a failure of experience transfer. The number of instances where MPI achieves a higher performance mean (than the baselines) is comparable to other classes, but a larger part of them lack statistical significance. We attribute this observation to the nature of the CAO problem itself: in compiler argument optimization, numerous distinct solutions can produce the same objective value (i.e., binary file size). Consequently, even with a higher-quality initial population, the final optimization results may not be statistically different. Therefore, MPI exhibits a strong ability to transfer experience to instances from unseen binary problem classes, which significantly broadens its applicability.

\begin{table}[htbp]
    \centering
    \caption{Overview of the Time Cost During Initialization. The Data Excludes the Time Cost of Function Evaluations and Only Presents the Methods' Intrinsic Costs.}
    \begin{tabular}{ccccc}
        \toprule[1pt]
        Rand           & OBL            & SVM-SS      & KAES        & MPI         \\
        \midrule
        \(\approx 0s\) & \(\approx 0s\) & \(\leq 1s\) & \(\leq 1s\) & \(3-4\) mins \\
        \bottomrule[1pt]
    \end{tabular}
    \label{tab: runtime}
\end{table}

\subsubsection{Computational Overhead and Practical Applicability}

In addition to solution quality, the computational overhead of the initialization phase is analyzed, with results presented in Table~\ref{tab: runtime}. Excluding the time for function evaluations, MPI requires 3$\sim$4 minutes, substantially longer than the baselines, which typically complete in under a second. This overhead suggests that MPI's practical advantage is most pronounced in scenarios with time-consuming function evaluations, where its significant performance gains can justify the initial time investment. Furthermore, it should be noted that for application scenarios with strict runtime constraints, MPI would not be a suitable choice. Future works will explore improvemnet of EMPI to reduce this time consumption. Specifically, given that approximately 99\% of MPI's runtime is consumed by the VAE model's fine-tuning during the experience transfer stage, accelerating this fine-tuning process will be a primary focus of future research.

\begin{figure*}[htbp]
    \centering
    \subfloat[]{\includegraphics[width=0.242\textwidth]{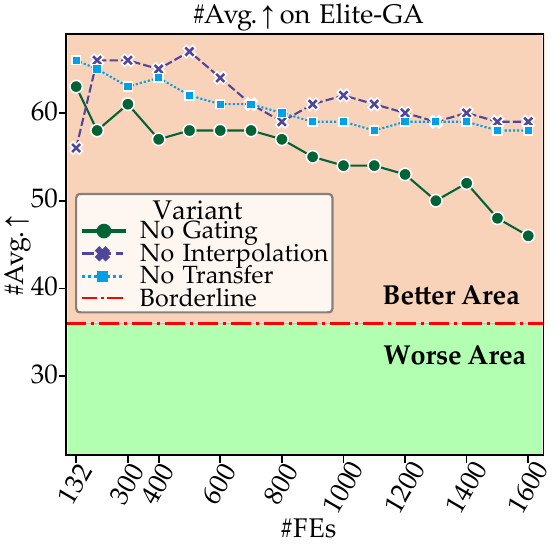}
    \label{fig: gatenet inter elitega avg}}
    \subfloat[]{\includegraphics[width=0.242\textwidth]{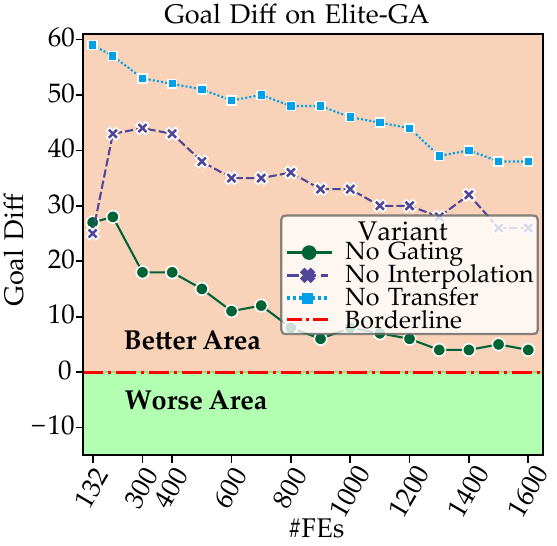}
    \label{fig: gatenet inter elitega goal diff}}
    \subfloat[]{\includegraphics[width=0.242\textwidth]{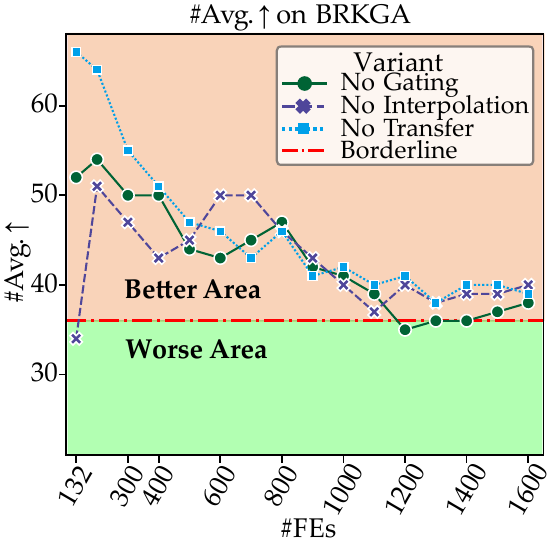}
    \label{fig: gatenet inter brkga avg}}
    \subfloat[]{\includegraphics[width=0.242\textwidth]{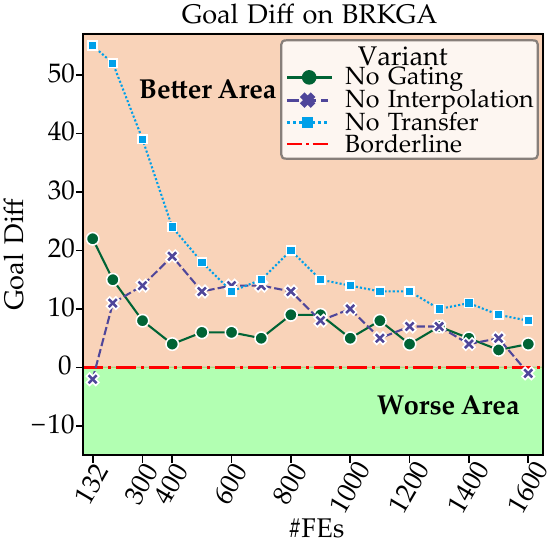}
    \label{fig: gatenet inter brkga goal diff}}
    \caption{Performance comparison of \textbf{MPI VS. its Variants} across \#FE budgets. The variants include \textbf{No Gating, No Interpolation, and No Transfer}. X-axis: \#FEs; Y-axis: performance metrics. \#Avg. \(\uparrow\): the number of instances where the MPI's average objective value of the best solution (from 30 evaluations) exceeds the Variants'. Goal Diff: Win number minus Loss number from statistical testing (W-D-L results) on best solutions over 30 evaluations. Each subplot includes a red horizontal line dividing the area into Better (MPI outperforms the Variants) and Worse (the Variants outperforms MPI) areas. 
    The subfigures present: (a) \#Avg. \(\uparrow\) metric on Elite-GA; (b) Goal Diff metric on Elite-GA; (c) \#Avg. \(\uparrow\) metric on BRKGA; (d) Goal Diff metric on BRKGA.}
    \label{fig: gatenet inter}
\end{figure*}

\subsection{Effectiveness of the Components in MPI}
\label{subsec: exp_gatenet}

To validate the contributions of MPI's key components, MPI was compared against two variants (No-Gating, and No-Transfer), which are defined as follows:
\begin{enumerate}
\item \textbf{No-Gating}: The gating network is removed. Experiences are selected randomly instead of being guided. The PI cost remains 132 FEs.
\item \textbf{No-Transfer}: The entire experience transfer mechanism is replaced with random sampling, followed by the interpolation step. The PI cost is 132 FEs.
\end{enumerate}
Additionally, to validate the effectiveness of the interpolation operator, the third variant, \textbf{No-Interpolation}, was included, which removes this operator and consumes only 112 FEs.

The performance comparison, shown in Fig.~\ref{fig: gatenet inter}, confirms that all ablated components are critical to MPI's performance, as MPI consistently outperforms all three variants across nearly all scenarios. The most significant performance degradation is observed in the No-Transfer variant, highlighting that the experience transfer mechanism is the cornerstone of MPI's effectiveness. The No-Gating variant also shows a substantial performance drop, which confirms the gating network's crucial role in selecting relevant experiences for the target problem. 

Interestingly, the No-Interpolation variant briefly outperforms MPI at the very beginning of the search when using BRKGA. This suggests that BRKGA's native operators are effective enough for an initial improvement. However, its performance quickly falls behind, indicating that the interpolation operator is superior for enhancing long-term population quality and is a vital contributor to MPI's overall efficacy in \#FEs constrained environments.

\begin{figure*}[htbp]
    \centering
    \subfloat[]{\includegraphics[width=0.242\textwidth]{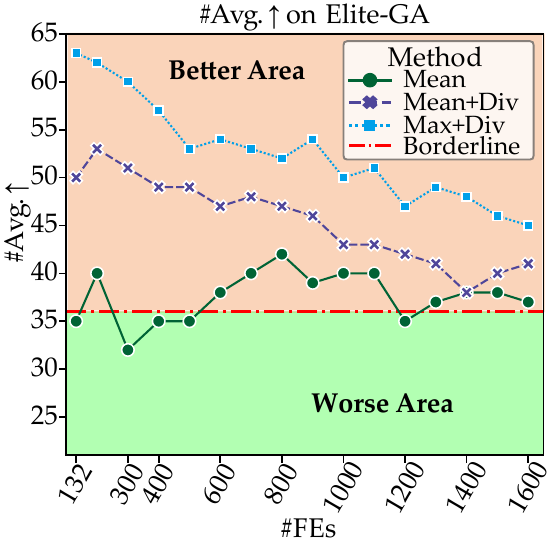}
    \label{fig: gatenet obj elitega avg}}
    \subfloat[]{\includegraphics[width=0.242\textwidth]{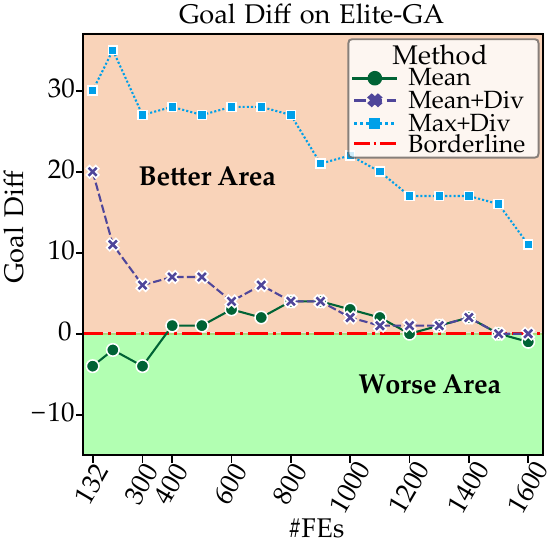}
    \label{fig: gatenet obj elitega goal diff}}
    \subfloat[]{\includegraphics[width=0.242\textwidth]{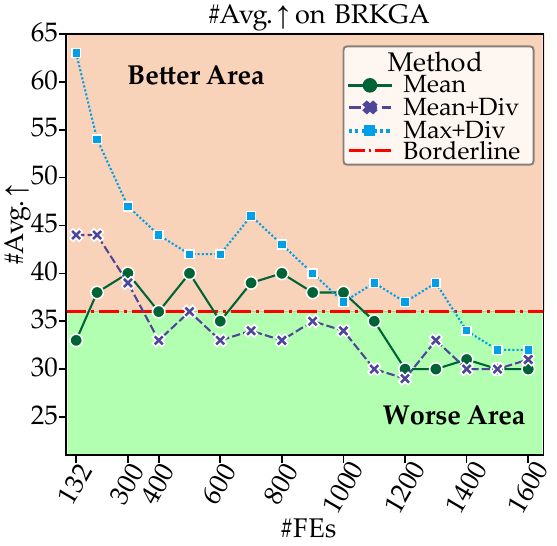}
    \label{fig: gatenet obj brkga avg}}
    \subfloat[]{\includegraphics[width=0.242\textwidth]{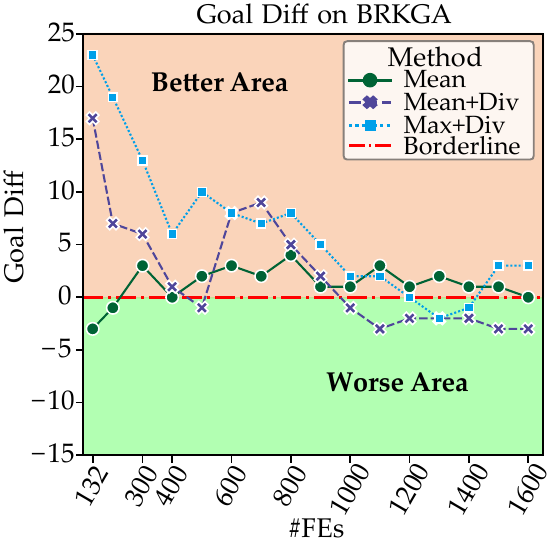}
    \label{fig: gatenet obj brkga goal diff}}
    \caption{Performance comparison of MPI using \textbf{Max as the gating network objective function VS. using Mean, Mean+Diversity, and Max+Diversity as objective functions} across \#FE budgets. X-axis: \#FEs; Y-axis: performance metrics. \#Avg. \(\uparrow\): the number of instances where the MPI with Max's average objective value of the best solution (from 30 evaluations) exceeds the other objective function's. Goal Diff: Win number minus Loss number from statistical testing (W-D-L results) on best solutions over 30 evaluations. Each subplot includes a red horizontal line dividing the area into Better (MPI with Max outperforms the the other objective function) and Worse (MPI with the other objective function outperforms Max) areas.
    The subfigures present: (a) \#Avg. \(\uparrow\) metric on Elite-GA; (b) Goal Diff metric on Elite-GA; (c) \#Avg. \(\uparrow\) metric on BRKGA; (d) Goal Diff metric on BRKGA.}
    \label{fig: gatenet obj}
\end{figure*}

\subsection{Choice of the Objective Function of Gating Network}
\label{subsec: exp_obj}

To justify the choice of the objective function used for training the gating network, four different functions were evaluated: Max, Mean, Max+Diversity, and Mean+Diversity. The Max function was ultimately adopted in the final MPI model. The formulations for these four objective functions are as follows:
\begin{enumerate}
    \item Max:
    \begin{equation}
        \max_{\mathbf{w}^G}\sum_{s\in S_G}\max_{\mathbf{x}\in X_s^{gen}}f_s\left(\mathbf{x}\right).
    \end{equation}
    \item Mean:
    \begin{equation}
        \max_{\mathbf{w}^G}\sum_{s\in S_G}\dfrac{1}{\left\vert X_s^{gen}\right\vert}\sum_{\mathbf{x}\in X_s^{gen}}f_s\left(\mathbf{x}\right).
    \end{equation}
    \item Max+Diversity:
    \begin{equation}
        \max_{\mathbf{w}^G}\sum_{s\in S_G}\left(\max_{\mathbf{x}\in X_s^{gen}}f_s\left(\mathbf{x}\right)
        +\mathrm{Diversity}\left(X_s^{gen}\right)\right).
    \end{equation}
    \item Mean+Diversity
    \begin{equation}
        \max_{\mathbf{w}^G}\sum_{s\in S_G}\left(\sum_{\mathbf{x}\in X_s^{gen}}f_s\left(\mathbf{x}\right)
        +\mathrm{Diversity}\left(X_s^{gen}\right)\right).
    \end{equation}
\end{enumerate}
Here, \(S_G\) is the additionally generated set of instances used to train the gating network, which also includes only instances of the three problem classes: OM, MC, and KP. \(f_s\) is the objective function of instance \(s\), and \(X_s^{gen}\) indicates the set of high-quality solutions generated for \(s\) using the weight parameters \(\mathbf{w}^G\) of the gating network. The detailed generation process is detailed in Sections IV-B and IV-C. To assess the diversity of the solution set \(X_s^{gen}\), a diversity function is utilized, which computes the mean of the average pairwise Hamming distances among all solutions in the set, as expressed by the following equation:
\begin{equation}
    \mathrm{Diversity}\left(X_s^{gen}\right) = \dfrac{1}{\left\vert X_s^{gen}\right\vert^2 d_s}\sum_{\mathbf{x}_1\in X_s^{gen}}\sum_{\mathbf{x}_2\in X_s^{gen}}\left\Vert\mathbf{x}_1-\mathbf{x}_2\right\Vert_1,
\end{equation}
where \(d_s\) is the dimensionality of \(s\). Moreover, since the range of the objective function varies across different instances, it is necessary to normalize \(f_s\left(\mathbf{x}\right)\), according to Eq.~(\ref{eq: fitness_normalized}).
After normalization, the vast majority of \(f_s\left(\mathbf{x}\right)\) values fall within the interval \(\left[0,1\right]\), with only a small portion slightly less than 0 or slightly greater than 1. Meanwhile, the range of \(\mathrm{Diversity}\left(X_s^{gen}\right)\) is also \(\left[0, 1\right]\), which ensures that when both the objective value metric and the diversity metric are considered together, the optimization process does not overly favor either one individually.

The results are presented in Fig~\ref{fig: gatenet obj}. At first glance, selecting the Max objective, which omits a diversity metric, may seem to diverge from common practices in population-based optimization. However, while no objective function dominates across the entire budget range (\(\left[132, 1600\right]\)), the choice of Max is strongly justified by MPI's focus on \#FE-constrained scenarios. As the figure illustrates, within the critical budget range of 300 to 1000 FEs, using the Max objective yields consistently superior performance. This is because, in early generations, prioritizing the single best solution (Max) provides a more direct and potent search direction. Although objectives incorporating diversity (e.g., Mean+Diversity) may become more advantageous for longer runs with ample \#FEs (\(>\)1200), Max proves to be the optimal choice for achieving rapid, high-quality results under the limited computational budgets that MPI is designed for.

\section{Conclusion and Discussion}
\label{sec:conclusion}

This work introduces MPI, a general-purpose PI method for binary optimization problems.
The core innovation of MPI lies in its problem-agnostic mechanism for representing, selecting, and transferring solving experiences.
By leveraging knowledge from diverse, previously solved problem instances, MPI constructs high-quality initial populations for any new problem instance.
This approach significantly accelerates the search process of GAs, especially under strict \#FEs budgets. 
Extensive experiments confirm that MPI substantially outperforms existing methods and can effectively transfer experiences to novel problem instances from unseen problems classes and are of higher dimensionality. 

Future work will focus on extending MPI's capabilities and enhancing its efficiency.
A promising direction is to evolve MPI from a static initializer into a dynamic guidance mechanism, where experiences are transferred adaptively throughout the search process.
This would sustain the benefits of transfer learning in longer optimization runs where the influence of the initial population diminishes.
Concurrently, addressing the significant computational overhead of the experience transfer stage is important. 
Developing methods to accelerate the model fine-tuning process will be crucial for broadening MPI's applicability to problems where the cost of function evaluations is not prohibitively high.
Furthermore, beyond the application of MPI for GA initialization, extending this adaptive experience transfer technique to neural optimizers~\cite{liu2023howgood} also represents a highly promising direction for future research.


\appendices
\section{Using PGPE in the Gating Network Optimization}
\label{append: pgpe}

\begin{algorithm}[htbp]
	\small
	\caption{PGPE in the Gating Network Optimization}
	\label{alg:pgpe optimizer}
	\SetKwInput{KwData}{Input}
	\SetKwInput{KwResult}{Output}
	\KwData{Gating network \(G\), instance set \(S_G\), objective function of the gating network \(g\left(s_G, G, \mathbf{w}^G\right)\).}
	\KwResult{Optimized weight parameters \(\mathbf{w}^G\) of \(G\).}
	\SetKw{To}{to}
	\SetKw{Append}{append}
	\SetKw{Into}{into}
	\SetKw{Break}{break}
	\SetKwProg{Fn}{Function}{:}{end}
	\SetKwFunction{Sort}{sort}
	\SetKwFunction{Normalize}{normalize}
	\SetKwFunction{Distance}{distance}
	\SetKwFunction{MSE}{MSE}
	\SetKwFunction{MLP}{MLP}
	\SetKwFunction{InsFeature}{ins\_feature}
	\SetKwComment{EmptyLine}{ }{ }
	\SetNoFillComment
	Initialize PGPE's parameters \(\boldsymbol{\sigma}^{init}\) (initial standard deviation vector), \(\alpha_{\mu}\) (learning rate of mean value), \(\alpha_{\sigma}\) (learning rate of standard deviation), \(\sigma^{limit}\) (lower limitation of standard deviation);\\
	\(\boldsymbol{\mu}, \boldsymbol{\sigma} \leftarrow\boldsymbol{0}, \boldsymbol{\sigma}^{init}\)\;
	\(\mathbf{w}^\prime, f^\prime\leftarrow \boldsymbol{\mu}, g\left(S_G, G, \boldsymbol{\mu}\right)\)\;
	\For{\(iter\leftarrow 1\) \To \(MaxIter\) }{
	\(\mathbf{w}_1,\mathbf{w}_2,\cdots,\mathbf{w}_{N}\leftarrow\) sampling \(N\) weights from \(\mathcal{N}\left(\boldsymbol{\mu},\boldsymbol{\sigma}\right)\)\;
	\(\mathbf{w}_{N+i}\leftarrow 2\boldsymbol{\mu}-\mathbf{w}_i\), where \(i=1,2,\cdots,N\)\;
	\(\boldsymbol{\epsilon}_1,\boldsymbol{\epsilon}_2,\cdots,\boldsymbol{\epsilon}_{N}\leftarrow \mathbf{w}_1-\boldsymbol{\mu},\mathbf{w}_2-\boldsymbol{\mu},\cdots,\mathbf{w}_{N}-\boldsymbol{\mu}\)\;
	\(\boldsymbol{\mu},\boldsymbol{\sigma},\mathbf{w}_i,\boldsymbol{\epsilon}_i\) are \(d\) dimension vector\;
	\(f_i\leftarrow g\left(S_G, G, \mathbf{w}_i\right)\), where \(i=1,2,\cdots,N\)\;
	\(f_b\leftarrow g\left(S_G, G, \boldsymbol{\mu}\right)\)\;
	\(\mathbf{w}_{\star},f_{\star}\leftarrow\) the best weight in \(\left\{\mathbf{w}_1, \mathbf{w}_2,\cdots,\mathbf{w}_{2N}, \boldsymbol{\mu}\right\}\)\;
	\lIf{\(f_{\star}\leq f^\prime\)}{\(\mathbf{w}^\prime, f^\prime\leftarrow \mathbf{w}_{\star},f_{\star}\)}
	\(\mathbf{M}\leftarrow\) a \(N\times d\) matrix, and \(\mathbf{M}_{ij}=\boldsymbol{\epsilon}_i^{\left(j\right)}\)\;
	\(\mathbf{f}^{M}\leftarrow\left[f_1-f_{N+1},f_2-f_{N+2},\cdots,f_{N}-f_{2N}\right]\)\;
	\(\mathbf{S}\leftarrow\) a \(N\times d\) matrix, and \(\mathbf{S}_{ij}=\frac{\left(\boldsymbol{\epsilon}_i^{\left(j\right)}\right)^2-\boldsymbol{\sigma}_i^2}{\boldsymbol{\sigma}_i}\)\;
	\resizebox{0.82\linewidth}{!}{\(\mathbf{f}^{S}\leftarrow\left[\frac{f_1+f_{N+1}}{2}-f_b, \frac{f_2+f_{N+2}}{2}-f_b,\cdots, \frac{f_N+f_{2N}}{2}-f_b\right]\)}\;

	\(\boldsymbol{\mu}, \boldsymbol{\sigma}\leftarrow \boldsymbol{\mu}+\alpha_{\mu}\mathbf{M}\mathbf{f}^{M}, \left\lfloor\boldsymbol{\sigma}+\alpha_{\sigma}\mathbf{S}\mathbf{f}^{S}\right\rfloor_{\sigma^{limit}}\)\;
	}
	\Return{\(\mathbf{w}^\prime\)}
\end{algorithm}

To optimize the weight parameters of the gating network \(G\) which is used to select suitable experiences, PGPE~\cite{sehnke_parameter-exploring_2010} is used as the optimizer.
The details are shown in Alg.~\ref{alg:pgpe optimizer}.
Specifically, PGPE employs the symmetric sampling exploration strategy (lines~6-9) and the strategy update method (lines~14-18).
It uses an iteratively updated multivariate Gaussian distribution to explore the vector space of the weight parameter.
The hyper-parameters in Alg.~\ref{alg:pgpe optimizer} are set to \(\boldsymbol{\sigma}^{init}=\boldsymbol{0.1}\), \(\alpha_{\mu}=0.01\), \(\alpha_{\sigma}=0.2\), and \(\sigma^{limit}=0.01\).

\begin{algorithm}[htbp]
	\small
	\caption{Solution Partitioning Algorithm}
	\label{alg:partition}
	\KwIn{The solution set\(X=\left[\mathbf{x}_1,\mathbf{x}_2,\cdots,\mathbf{x}_n\right]\), fitness vector \(\mathbf{y}=\left[y_1,y_2,\cdots,y_n\right]\), the number of subsets \(e^l\)}
	\KwOut{Partitioned subsets \(X_1,X_2,\cdots,X_{e^l}\)}
	\(Gr\leftarrow\) An empty Map\;
	\For{\(i\leftarrow 1,2,\cdots, n\)}{
		\lIf{\(Gr\left[y_i\right]\) not exist}{\(Gr\left[y_i\right]\leftarrow \emptyset\)}
		Append \(\mathbf{x}_i\) into \(Gr\left[y_i\right]\)\;
	}
	\(\mathbf{f}\leftarrow\) The descending order keys of \(Gr\)\;
	\(i, m\leftarrow 1, 0\)\;
	\(X_1,X_2,\cdots,X_{e^l}\leftarrow\emptyset,\emptyset,\cdots,\emptyset\)\;
	\For{\(j\leftarrow 1,2,\cdots,\left\vert \mathbf{f}\right\vert\)}{
		\(c_1\leftarrow\left(e^l-j<\left\vert \mathbf{f}\right\vert-i\right)\wedge\left(i<\left\vert \mathbf{f}\right\vert\right)\)\;
		\(c_2\leftarrow\left(\frac{\left\vert \mathbf{f}\right\vert\times i}{e^l}-m\leq\frac{2\times\left\vert Gr\left[f_j\right]\right\vert}{3}\right)\wedge \left(i<e^l\right)\)\;
		\uIf{\(\left(X_i=\emptyset\right) \vee \left(\neg\left(c_1\vee c_2\right)\right)\)}{
			\(X_i\leftarrow X_i\cup Gr\left[f_j\right]\)\;
			\(m\leftarrow m+\left\vert Gr\left[f_j\right]\right\vert\)\;
		}\lElse{
			\(i\leftarrow i+1\)
		}
	}
    \Return{\(X_1,X_2,\cdots,X_{e^l}\)}
\end{algorithm}

\section{Solutions Partitioning by Fitness Ranking}
\label{append: solution_divides}

This partitioning strategy classifies a given set of solutions \(X\) into a predefined \(e^l\) subsets, based on their objective values \(\mathbf{y}\). The objective is to distribute the solutions as uniformly as possible across the subsets while preserving a descending order of objective value. A key constraint is that all solutions with identical objective values must be assigned to the same subset.

As shown in Alg.~\ref{alg:partition}, the process begins by consolidating solutions into fitness groups, each containing all solutions with the same objective value (lines~1-5). These groups, which are the atomic units for partitioning, are then sorted in descending order based on their objective value.

The algorithm's core involves allocating these sorted fitness groups into the \(e^l\) subsets. To guide this, an ideal average number of solutions per subset \(\frac{\left\vert X\right\vert}{e^l}\) is calculated. The allocation proceeds iteratively with objective values from high to low, assigning groups to the current subset until a balancing criterion is met (lines~12-13). The algorithm advances to the next subset when either the current subset's size is sufficiently close to the target average or when it is necessary to ensure that the remaining groups can fill subsequent empty subsets (line~16). This dual-condition approach promotes a uniform distribution and prevents any subset from becoming disproportionately large or empty (lines~10-12). The final result is a list of \(e^l\) subsets, ordered from highest to lowest objective value.

\begin{algorithm}[htbp]
	\small
	\caption{Interpolation Operator}
	\label{alg:interpolation}
	\KwIn{The solution set\(X\in\left\{0,1\right\}^{n\times d}\), objective value vector \(\mathbf{y}\in\mathbb{R}^n\), the number of generation \(q_m\)}
	\KwOut{New solution set \(X\in\left\{0,1\right\}^{\left(n+q_m\right)\times d}\) and its objective value vector \(\mathbf{y}\in\mathbb{R}^{n+q_m}\)}
	\(X_e \leftarrow \) Top 10\% of solutions in \(X\) based on \(\mathbf{y}\)\;
	\(X_m \leftarrow X \setminus X_e\)\;
	\For{\(\text{iter}\leftarrow 1, \dots, q_m\)}{
	\(P_e \leftarrow \) Randomly select 2 solutions from \(X_e\)\;
	\(P_m \leftarrow \) Randomly select 2 solutions from \(X_m\)\;
	\(X_p \leftarrow P_e \cup P_m\)\;
    Initialize a new solution \(\mathbf{x}_{new}\)\;
    \For{\(i \leftarrow 1, \dots, d\)}{
        \uIf{all solutions in \(X_p\) have the same value at dimension \(i\)}{
            \(\mathbf{x}_{new}[i] \leftarrow X_p[1][i]\)\;
        }
        \Else{
            \(r \sim U(0, 1)\)\;
            \uIf{\(r < \text{average value of dimension } i \text{ in } X_p\)}{
                \(\mathbf{x}_{new}[i]\leftarrow 1\)
            }
            \lElse{\(\mathbf{x}_{new}[i]\leftarrow 0\)}
        }
    }
    \(y_{new} \leftarrow f_s(\mathbf{x}_{new})\)\;
    \(X,\mathbf{y} \leftarrow X \cup \{\mathbf{x}_{new}\}, \mathbf{y} \cup \{y_{new}\}\)\;
	}
    Sort \(X\) and \(\mathbf{y}\) in descending order based on values in \(\mathbf{y}\)\;
    \Return{\(X, \mathbf{y}\)}
\end{algorithm}

\section{Interpolation-based Solution Generation Operator}
\label{append: interpolation}

The interpolation-based solution generation operator is designed to produce new candidate solutions for binary optimization problems by leveraging information from an existing solution set. Given an initial solution set \(X\) and their corresponding objective values \(\mathbf{y}\), the operator generates new solutions and integrates them into the solution set.

The process begins by partitioning the current solution set \(X\). Based on their objective values in \(\mathbf{y}\), the top 10\% of solutions are designated as the elite set (\(X_{elite}\)), while the remaining 90\% form the mediocre set (\(X_m\)).
The core of the operator is repeated generation process that runs for a predefined number of times, denoted as \(q_m\). In each iteration, a group of four parent solutions is selected: two are chosen randomly from \(X_{elite}\) and the other two from \(X_m\). A new binary solution vector, \(\mathbf{x}_{new}\), is then constructed dimension by dimension. For each dimension \(i\), if all four parent solutions share the same value, \(\mathbf{x}_{new}\) inherits this value. Otherwise, the average value of the parents' \(i\)-th dimension, denoted as \(avg\), is calculated. A random number \(r\) is sampled uniformly at random from \(\left[0, 1\right]\). The \(i\)-th dimension of \(\mathbf{x}_{new}\) is set to 1 if \(r < avg\), and 0 otherwise.
Once the new solution \(\mathbf{x}_{new}\) is fully constructed, its objective value \(y_{{new}}\) is evaluated on the given problem instance \(s\), such that \(y_{{new}} = f_s(\mathbf{x}_{new})\). Both the new solution and its objective value are then added to the sets \(X\) and \(\mathbf{y}\), respectively.

After \(q_m\) repetitions, the entire solution set \(X\) is re-sorted in descending order based on the updated objective values in \(\mathbf{y}\). The complete procedure is detailed in Alg.~\ref{alg:interpolation}.

\bibliographystyle{IEEEtran}
\bibliography{EMPI}

@incollection{abdul_halim_algorithm_2024,
  title     = {Algorithm Initialization: Categories and Assessment},
  author    = {Abdul Halim, Abdul Hanif and Das, Swagatam and Ismail, Idris},
  booktitle = {Into a Deeper Understanding of Evolutionary Computing: Exploration, Exploitation, and Parameter Control: Volume 1},
  pages     = {1--100},
  year      = {2024},
  publisher = {Springer}
}

@inproceedings{eshelman_crossover_1991,
  author    = {Eshelman, L},
  booktitle = {Proceedings of ICGA 1991, San Diego, CA},
  title     = {{O}n crossover as an evolutionarily viable strategy}
}

@article{feng_autoencoding_2017,
  author  = {Liang Feng and
             Yew{-}Soon Ong and
             Siwei Jiang and
             Abhishek Gupta},
  title   = {Autoencoding Evolutionary Search With Learning Across Heterogeneous Problems},
  journal = {{IEEE} Trans. Evol. Comput.},
  volume  = {21},
  number  = {5},
  pages   = {760--772},
  year    = {2017}
}

@article{feng_memes_2015,
  author  = {Liang Feng and
             Yew{-}Soon Ong and
             Ah{-}Hwee Tan and
             Ivor W. Tsang},
  title   = {Memes as building blocks: a case study on evolutionary optimization + transfer learning for routing problems},
  journal = {Memetic Comput.},
  volume  = {7},
  number  = {3},
  pages   = {159--180},
  year    = {2015}
}

@article{feng_memetic_2015,
  author  = {Liang Feng and
             Yew{-}Soon Ong and
             Meng{-}Hiot Lim and
             Ivor W. Tsang},
  title   = {Memetic Search With Interdomain Learning: {A} Realization Between {CVRP} and {CARP}},
  journal = {{IEEE} Trans. Evol. Comput.},
  volume  = {19},
  number  = {5},
  pages   = {644--658},
  year    = {2015}
}

@article{feng_towards_2022,
  author  = {Liang Feng and
             Yuxiao Huang and
             Ivor W. Tsang and
             Abhishek Gupta and
             Ke Tang and
             Kay Chen Tan and
             Yew{-}Soon Ong},
  title   = {Towards Faster Vehicle Routing by Transferring Knowledge From Customer Representation},
  journal = {{IEEE} Trans. Intell. Transp. Syst.},
  volume  = {23},
  number  = {2},
  pages   = {952--965},
  year    = {2022}
}

@book{garey_computers_1979,
  author    = {M. R. Garey and
               David S. Johnson},
  isbn      = {0-7167-1044-7},
  publisher = {W. H. Freeman},
  title     = {Computers and Intractability: A Guide to the Theory of NP-Completeness},
  year      = {1979}
}

@inproceedings{gong_effects_2023,
  author    = {Cheng Gong and
               Lie Meng Pang and
               Yang Nan and
               Hisao Ishibuchi and
               Qingfu Zhang},
  title     = {Effects of Initialization Methods on the Performance of Multi-Objective Evolutionary Algorithms},
  booktitle = {Proceedings of {IEEE SMC} 2023, Honolulu, HI},
  pages     = {1168--1175}
}

@inproceedings{gonzalez-duque_survey_2024,
  author    = {Miguel Gonz{\'{a}}lez Duque and
               Richard Michael and
               Simon Bartels and
               Yevgen Zainchkovskyy and
               S{\o}ren Hauberg and
               Wouter Boomsma},
  title     = {A survey and benchmark of high-dimensional Bayesian optimization of
               discrete sequences},
  booktitle = {Proceedings of {NeurIPS} 2024, BC, Canada},
  pages     = {140478-140508}
}

@article{hao-evaluated_2025,
  author  = {Hao Hao and
             Xiaoqun Zhang and
             Aimin Zhou},
  title   = {Un-evaluated solutions may be valuable in expensive optimization},
  journal = {Swarm Evol. Comput.},
  volume  = {94},
  pages   = {101905},
  year    = {2025}
}

@inproceedings{hu_contamination_2010,
  author    = {Yingjie Hu and
               Jianqiang Hu and
               Yifan Xu and
               Fengchun Wang and
               Rongzeng Cao},
  booktitle = {Proceedings of {WSC} 2010, MD, USA},
  pages     = {2678--2681},
  title     = {Contamination control in food supply chain}
}

@incollection{iman_latin_2008,
  author    = {Iman, Ronald L.},
  publisher = {John Wiley \& Sons, Ltd},
  title     = {{L}atin {H}ypercube {S}ampling},
  booktitle = {Encyclopedia of Quantitative Risk Analysis and Assessment},
  year      = {2008}
}

@article{jiang_fast_2021,
  author  = {Min Jiang and
             Zhenzhong Wang and
             Liming Qiu and
             Shihui Guo and
             Xing Gao and
             Kay Chen Tan},
  title   = {A Fast Dynamic Evolutionary Multiobjective Algorithm via Manifold Transfer Learning},
  journal = {{IEEE} Trans. Cybern.},
  volume  = {51},
  number  = {7},
  pages   = {3417--3428},
  year    = {2021}
}

@article{jiang_smartest_2022,
  author  = {He Jiang and
             Guojun Gao and
             Zhilei Ren and
             Xin Chen and
             Zhide Zhou},
  title   = {{SMARTEST:} {A} Surrogate-Assisted Memetic Algorithm for Code Size
             Reduction},
  journal = {{IEEE} Trans. Reliab.},
  volume  = {71},
  number  = {1},
  pages   = {190--203},
  year    = {2022}
}

@inproceedings{juarez_recombination_2024,
  author    = {Julio Ju{\'a}rez and
               Carlos A. Brizuela and
               Hugo Terashima-Mar{\'\i}n and
               Carlos A. Coello Coello},
  booktitle = {Proceedings of {CEC} 2024, Yokohama, Japan},
  pages     = {1--8},
  title     = {Recombination Operators for the Multi-Objective Team Formation Problem in Social Networks}
}

@inproceedings{keedwell_novel_2018,
  author    = {Edward C. Keedwell and
               Mathieu Br{\'{e}}villiers and
               Lhassane Idoumghar and
               Julien Lepagnot and
               Hojjat Rakhshani},
  title     = {A Novel Population Initialization Method Based on Support Vector Machine},
  booktitle = {Proceedings of {IEEE SMC} 2018, Miyazaki, Japan},
  pages     = {751--756}
}

@inproceedings{kingma_auto-encoding_2014,
  author    = {Kingma, Diederik P. and Welling, Max},
  booktitle = {Proceedings of {ICLR} 2014, Banff, AB, Canada},
  title     = {Auto-Encoding Variational Bayes.},
  year      = {2014}
}

@article{lan_region-focused_2022,
  author  = {Wenxing Lan and
             Ziyuan Ye and
             Peijun Ruan and
             Jialin Liu and
             Peng Yang and
             Xin Yao},
  title   = {Region-Focused Memetic Algorithms With Smart Initialization for Real-World Large-Scale Waste Collection Problems},
  journal = {{IEEE} Trans. Evol. Comput.},
  volume  = {26},
  number  = {4},
  pages   = {704--718},
  year    = {2022}
}

@inproceedings{leskovec_predicting_2010,
  author    = {Jure Leskovec and
               Daniel P. Huttenlocher and
               Jon M. Kleinberg},
  booktitle = {Proceedings of {WWW} 2010, NC, USA},
  title     = {Predicting positive and negative links in online social networks}
}

@article{li_chance-constrained_2024,
  author  = {Xuanfeng Li and
             Shengcai Liu and
             Jin Wang and
             Xiao Chen and
             Yew{-}Soon Ong and
             Ke Tang},
  title   = {Chance-Constrained Multiple-Choice Knapsack Problem: Model, Algorithms, and Applications},
  journal = {{IEEE} Trans. Cybern.},
  volume  = {54},
  number  = {12},
  pages   = {7969--7980},
  year    = {2024}
}

@article{li_tailoring_2024,
  author  = {Yuanrui Li and
             Qiuhong Zhao and
             Shengxiang Yang and
             Yinan Guo},
  title   = {Tailoring Evolutionary Algorithms to Solve the Multiobjective Location-Routing Problem for Biomass Waste Collection},
  journal = {{IEEE} Trans. Evol. Comput.},
  volume  = {28},
  number  = {2},
  pages   = {489--500},
  year    = {2024}
}

@article{liu_evolutionary-multimodal_2024,
  author  = {Yiping Liu and
             Liting Xu and
             Yuyan Han and
             Xiangxiang Zeng and
             Gary G. Yen and
             Hisao Ishibuchi},
  journal = {{IEEE} Trans. Evol. Comput.},
  number  = {2},
  pages   = {516--530},
  title   = {Evolutionary Multimodal Multiobjective Optimization for Traveling Salesman Problems},
  volume  = {28},
  year    = {2024}
}

@article{londe_biased_2025,
  author  = {Mariana A. Londe and
             Luciana S. Pessoa and
             Carlos Eduardo de Andrade and
             Mauricio G. C. Resende},
  title   = {Biased random-key genetic algorithms: {A} review},
  journal = {Eur. J. Oper. Res.},
  volume  = {321},
  number  = {1},
  pages   = {1--22},
  year    = {2025}
}

@article{lu_competition_2015,
  author  = {Wei Lu and
             Wei Chen and
             Laks V. S. Lakshmanan},
  title   = {From Competition to Complementarity: Comparative Influence Diffusion and Maximization},
  journal = {Proc. {VLDB} Endow.},
  volume  = {9},
  number  = {2},
  pages   = {60--71},
  year    = {2015}
}

@book{martello_knapsack_1990,
  author    = {Martello, Silvano and Toth, Paolo},
  publisher = {John Wiley \& Sons, Inc.},
  title     = {Knapsack problems: algorithms and computer implementations},
  year      = {1990}
}

@inproceedings{mcauley_learning_2012,
  author    = {Julian J. McAuley and
               Jure Leskovec},
  booktitle = {Proceedings of {NIPS} 2012, NV, USA},
  title     = {Learning to Discover Social Circles in Ego Networks}
}

@book{mitchell_introduction_1998,
  author    = {Mitchell, Melanie},
  isbn      = {978-0-262-63185-3},
  publisher = {MIT Press},
  title     = {An Introduction to Genetic Algorithms},
  year      = {1998}
}

@inproceedings{oh_combinatorial_2019,
  author    = {ChangYong Oh and
               Jakub M. Tomczak and
               Efstratios Gavves and
               Max Welling},
  booktitle = {Proceedings of {NeurIPS} 2019, BC, Canada},
  pages     = {2910--2920},
  title     = {Combinatorial Bayesian Optimization using the Graph Cartesian Product}
}

@article{omidvar_review_2022,
  author  = {Mohammad Nabi Omidvar and
             Xiaodong Li and
             Xin Yao},
  title   = {A Review of Population-Based Metaheuristics for Large-Scale Black-Box Global Optimization - Part {II}},
  journal = {{IEEE} Trans. Evol. Comput.},
  volume  = {26},
  number  = {5},
  pages   = {823--843},
  year    = {2022}
}

@article{rahnamayan_opposition-based_2008,
  author  = {Shahryar Rahnamayan and
             Hamid R. Tizhoosh and
             Magdy M. A. Salama},
  title   = {Opposition-Based Differential Evolution},
  journal = {{IEEE} Trans. Evol. Comput.},
  volume  = {12},
  number  = {1},
  pages   = {64--79},
  year    = {2008}
}

@article{segredo_comparison_2018,
  author  = {Eduardo Segredo and
             Ben Paechter and
             Carlos Segura and
             Carlos Ignacio Gonzalez{-}Vila},
  title   = {On the comparison of initialisation strategies in differential evolution for large scale optimisation},
  journal = {Optim. Lett.},
  volume  = {12},
  number  = {1},
  pages   = {221--234},
  year    = {2018}
}

@article{sehnke_parameter-exploring_2010,
  author  = {Frank Sehnke and
             Christian Osendorfer and
             Thomas R{\"u}ckstie{\ss} and
             Alex Graves and
             Jan Peters and
             J{\"u}rgen Schmidhuber},
  journal = {Neural Netw.},
  number  = {4},
  pages   = {551--559},
  title   = {Parameter-exploring policy gradients},
  volume  = {23},
  year    = {2010}
}

@article{song_kriging-assisted_2021,
  author  = {Zhenshou Song and
             Handing Wang and
             Cheng He and
             Yaochu Jin},
  title   = {A Kriging-Assisted Two-Archive Evolutionary Algorithm for Expensive Many-Objective Optimization},
  journal = {{IEEE} Trans. Evol. Comput.},
  volume  = {25},
  number  = {6},
  pages   = {1013--1027},
  year    = {2021}
}

@article{tan_evolutionary_2021,
  author  = {Kay Chen Tan and
             Liang Feng and
             Min Jiang},
  title   = {Evolutionary Transfer Optimization - {A} New Frontier in Evolutionary Computation Research},
  journal = {{IEEE} Comput. Intell. Mag.},
  volume  = {16},
  number  = {1},
  pages   = {22--33},
  year    = {2021}
}

@article{wei_distributed_2023,
  author  = {Feng{-}Feng Wei and
             Wei{-}Neng Chen and
             Qing Li and
             Sang{-}Woon Jeon and
             Jun Zhang},
  title   = {Distributed and Expensive Evolutionary Constrained Optimization With On-Demand Evaluation},
  journal = {{IEEE} Trans. Evol. Comput.},
  volume  = {27},
  number  = {3},
  pages   = {671--685},
  year    = {2023}
}

@book{yu_introduction_2012,
  author    = {Yu, Xinjie and Gen, Mitsuo},
  isbn      = {978-1-4471-2569-3},
  publisher = {Springer},
  title     = {Introduction to Evolutionary Algorithms},
  year      = {2012}
}

@article{zhang_surrogate-assisted_2021,
  author  = {Fangfang Zhang and
             Yi Mei and
             Su Nguyen and
             Mengjie Zhang and
             Kay Chen Tan},
  title   = {Surrogate-Assisted Evolutionary Multitask Genetic Programming for Dynamic Flexible Job Shop Scheduling},
  journal = {{IEEE} Trans. Evol. Comput.},
  volume  = {25},
  number  = {4},
  pages   = {651--665},
  year    = {2021}
}

@article{zhou_learnable_2021,
  author  = {Lei Zhou and
             Liang Feng and
             Abhishek Gupta and
             Yew{-}Soon Ong},
  title   = {Learnable Evolutionary Search Across Heterogeneous Problems via Kernelized Autoencoding},
  journal = {{IEEE} Trans. Evol. Comput.},
  volume  = {25},
  number  = {3},
  pages   = {567--581},
  year    = {2021}
}

@inproceedings{zhou_solving_2017,
  author    = {Lei Zhou and
               Liang Feng and
               Abhishek Gupta and
               Yew{-}Soon Ong and
               K. Liu and
               C. Chen and
               Edwin Hsing{-}Mean Sha and
               B. Yang and
               B. W. Yan},
  title     = {Solving dynamic vehicle routing problem via evolutionary search with learning capability},
  booktitle = {Proceedings of {CEC} 2017, San Sebasti{\'{a}}n, Spain},
  pages     = {890--896}
}

@article{xue_surrogate-assisted-search_2024,
  author  = {Xiaoming Xue and
             Yao Hu and
             Liang Feng and
             Kai Zhang and
             Linqi Song and
             Kay Chen Tan},
  title   = {Surrogate-Assisted Search with Competitive Knowledge Transfer for Expensive Optimization},
  journal = {{IEEE} Trans. Evol. Comput.},
  year    = {2024},
  volume  = {},
  number  = {},
  pages   = {1-1}
}

@inproceedings{kazimipour_review_2014,
  author    = {Borhan Kazimipour and
               Xiaodong Li and
               A. Kai Qin},
  title     = {A review of population initialization techniques for evolutionary algorithms},
  booktitle = {Proceedings of {CEC} 2014, Beijing, China},
  pages     = {2585--2592}
}

@article{blank_pymoo_2020,
  author       = {Julian Blank and
                  Kalyanmoy Deb},
  title        = {Pymoo: Multi-Objective Optimization in Python},
  journal      = {{IEEE} Access},
  volume       = {8},
  pages        = {89497--89509},
  year         = {2020}
}

@inproceedings{lepikhin_gshard_2021,
  author       = {Dmitry Lepikhin and
                  HyoukJoong Lee and
                  Yuanzhong Xu and
                  Dehao Chen and
                  Orhan Firat and
                  Yanping Huang and
                  Maxim Krikun and
                  Noam Shazeer and
                  Zhifeng Chen},
  title        = {GShard: Scaling Giant Models with Conditional Computation and Automatic
                  Sharding},
  booktitle    = {Proceedings of {ICLR} 2021, Virtual Event}
}

@article{liu2023howgood,
  author       = {Shengcai Liu and
                  Yu Zhang and
                  Ke Tang and
                  Xin Yao},
  title        = {How Good is Neural Combinatorial Optimization? {A} Systematic Evaluation
                  on the Traveling Salesman Problem},
  journal      = {{IEEE} Comput. Intell. Mag.},
  volume       = {18},
  number       = {3},
  pages        = {14--28},
  year         = {2023}
}

@article{dai2023saliencyattack,
  author       = {Zeyu Dai and
                  Shengcai Liu and
                  Qing Li and
                  Ke Tang},
  title        = {Saliency Attack: Towards Imperceptible Black-box Adversarial Attack},
  journal      = {{ACM} Trans. Intell. Syst. Technol.},
  volume       = {14},
  number       = {3},
  pages        = {45:1--45:20},
  year         = {2023}
}

@article{wang2025evolving,
  title     ={Evolving Generalizable Parallel Algorithm Portfolios for Binary Optimization Problems via Domain-Agnostic Instance Generation},
  author    ={Wang, Zhiyuan and Liu, Shengcai and Yang, Peng and Tang, Ke},
  journal   ={{IEEE} Trans. Evol. Comput.},
  volume    ={Early Access},
  year      ={2025}
}

@article{said2023featureselection,
  author       = {Rihab Said and
                  Maha Elarbi and
                  Slim Bechikh and
                  Carlos Artemio Coello Coello and
                  Lamjed Ben Said},
  title        = {Discretization-Based Feature Selection as a Bilevel Optimization Problem},
  journal      = {{IEEE} Trans. Evol. Comput.},
  volume       = {27},
  number       = {4},
  pages        = {893--907},
  year         = {2023}
}

\clearpage

\twocolumn[
  \centering
  {\huge\bfseries Supplementary for the Paper \\``A Novel Population Initialization Method via Adaptive Experience Transfer for General-Purpose Binary Evolutionary Optimization"}
  \vspace{2em}
]

\setcounter{table}{4}
\setcounter{section}{0}
\renewcommand{\appendixname}{Supplement}

\begin{table*}[htbp]
    \centering
    \caption{Mean Value and Std of MPI and Baselines on Each Instance over 30 Runs with GA-Elite. The Highest Average Performances are Highlighted in \textbf{black}. The Wilcoxon Rank-Sum Tests (\(p=0.05\)) is also Indicated by \(\uparrow\) \(\rightarrow\) and \(\downarrow\); \(\uparrow\): the Baseline Significantly Better, \(\rightarrow\): No Significant Difference, \(\uparrow\): MPI Significantly Better. \#W-D-L: Counts of \(\uparrow\) \(\rightarrow\) and \(\downarrow\). \# Highest Avg: the Number of Instances where the Method Has the Highest Average Performance.}
      \resizebox{0.75\textwidth}{!}{\begin{tabular}{cccccccc}
      \toprule[1pt]
      \makecell{Problem\\Class} & Dim & idx & MPI & Rand & SVM-SS & OBL & KAES\\
      \hhline{--------}
      \multirow{13}[0]{*}{MM} & \multirow{3}[0]{*}{40} & 0     & \textbf{37.3±1.22} & 31.933±1.59 ↓ & 35.033±1.64 ↓ & 32.8±1.14 ↓ & 38.433±0.989 ↑ \\
            &       & 1     & \textbf{37.333±1.19} & 32.6±1.89 ↓ & 34.9±1.56 ↓ & 33.467±1.43 ↓ & 36.967±1.05 → \\
            &       & 2     & \textbf{37.033±1.25} & 31.8±1.81 ↓ & 35.1±1.27 ↓ & 33.533±1.45 ↓ & 37.733±1.18 ↑ \\
            \hhline{~-------}
            & \multirow{3}[0]{*}{60} & 0     & \textbf{51.833±1.83} & 45.5±2.05 ↓ & 48.833±2.02 ↓ & 45.6±2.23 ↓ & 50.3±1.73 ↓ \\
            &       & 1     & \textbf{50.9±2.12} & 45.467±2.35 ↓ & 49.8±2.21 ↓ & 47±2.19 ↓ & 51.333±1.7 → \\
            &       & 2     & \textbf{52.9±1.99} & 45.133±2.69 ↓ & 48.967±2.07 ↓ & 46.9±2.2 ↓ & 50.667±2.3 ↓ \\
            \hhline{~-------}
            & \multirow{3}[0]{*}{80} & 0     & \textbf{66.567±2.26} & 57.467±2.62 ↓ & 63.1±2.56 ↓ & 60.333±2.52 ↓ & 65.067±1.75 ↓ \\
            &       & 1     & \textbf{66.3±1.62} & 56.6±2.29 ↓ & 62.733±1.98 ↓ & 58.067±2.56 ↓ & 64.467±2.11 ↓ \\
            &       & 2     & \textbf{65.733±2.42} & 57.4±2.65 ↓ & 62.6±1.76 ↓ & 59.333±2.27 ↓ & 65.867±1.84 → \\
            \hhline{~-------}
            & \multirow{3}[0]{*}{100} & 0     & \textbf{79.9±3.03} & 69.633±3.1 ↓ & 74.833±2.41 ↓ & 71.067±2.08 ↓ & 76.5±2.69 ↓ \\
            &       & 1     & \textbf{79.867±3.45} & 69.667±3.25 ↓ & 76.3±2.87 ↓ & 73.833±3.38 ↓ & 75.8±2.39 ↓ \\
            &       & 2     & \textbf{81.033±3.23} & 69.167±2.9 ↓ & 73.833±3.13 ↓ & 73.3±2.38 ↓ & 78.1±3.07 ↓ \\
            \hhline{~-------}
            & \multicolumn{2}{c}{\# W-D-L (VS. MPI)} &       & 0-0-12 & 0-0-12 & 0-0-12 & 2-3-7 \\
            \hhline{--------}
            \multirow{13}[0]{*}{MC} & \multirow{3}[0]{*}{40} & 0     & \textbf{213.33±4.14} & 207.2±3.22 ↓ & 210.9±3.74 ↓ & 208.93±3.45 ↓ & 210.67±3.71 ↓ \\
            &       & 1     & \textbf{300.63±2.66} & 297.2±3.53 ↓ & 300.47±3.63 → & 298.1±3.79 ↓ & 297.03±3.02 ↓ \\
            &       & 2     & 150.03±3.41 & 147.5±4.28 ↓ & 151.93±2.5 ↑ & 147.8±2.75 ↓ & 151.73±2.66 ↑ \\
            \hhline{~-------}
            & \multirow{3}[0]{*}{60} & 0     & 306.1±3.69 & 296.73±5.19 ↓ & 303.97±5.03 ↓ & 298.87±4.39 ↓ & 309.8±3.62 ↑ \\
            &       & 1     & \textbf{579.77±2.47} & 573.5±3.43 ↓ & 578.5±4.48 → & 574±3.9 ↓ & 577.3±4.45 ↓ \\
            &       & 2     & 461.03±3.54 & 457.7±4.85 ↓ & 463.83±5.68 → & 458.4±3.51 ↓ & 459.23±4.45 ↓ \\
            \hhline{~-------}
            & \multirow{3}[0]{*}{80} & 0     & 501.3±6.13 & 489.7±8.78 ↓ & 507.63±7.37 ↑ & 497.5±7.83 ↓ & 499.83±9.41 → \\
            &       & 1     & 559±6.66 & 547.6±7.84 ↓ & 559.33±5.26 → & 549.1±8.32 ↓ & 552.83±5.54 ↓ \\
            &       & 2     & 633.23±6.09 & 622.73±7.93 ↓ & 635.07±6.26 → & 626.73±6.88 ↓ & 632.33±5.04 → \\
            \hhline{~-------}
            & \multirow{3}[0]{*}{100} & 0     & \textbf{1218.4±7.86} & 1202.4±11 ↓ & 1212.9±10.5 ↓ & 1208.2±10.9 ↓ & 1208.2±5.7 ↓ \\
            &       & 1     & \textbf{1315.7±8.54} & 1299.3±9.15 ↓ & 1306.6±9.35 ↓ & 1301.9±8.72 ↓ & 1305.7±10 ↓ \\
            &       & 2     & \textbf{1380.9±8.16} & 1362.4±7.95 ↓ & 1375.6±8.58 ↓ & 1365.7±8.36 ↓ & 1369.8±10 ↓ \\
            \hhline{~-------}
            & \multicolumn{2}{c}{\# W-D-L (VS. MPI)} &       & 0-0-12 & 2-5-5 & 0-0-12 & 2-2-8 \\
            \hhline{--------}
            \multirow{13}[0]{*}{KP} & \multirow{3}[0]{*}{40} & 0     & \textbf{11.743±0.0974} & 11.605±0.0966 ↓ & 11.613±0.109 ↓ & 11.612±0.102 ↓ & 11.644±0.113 ↓ \\
            &       & 1     & 11.945±0.133 & 11.77±0.12 ↓ & 11.824±0.159 ↓ & 11.83±0.125 ↓ & 11.965±0.135 → \\
            &       & 2     & \textbf{11.057±0.0809} & 10.976±0.11 ↓ & 11.038±0.0745 → & 11.006±0.078 ↓ & 11.029±0.0875 → \\
            \hhline{~-------}
            & \multirow{3}[0]{*}{60} & 0     & \textbf{22.494±0.118} & 21.978±0.615 ↓ & 22.225±0.172 ↓ & 22.136±0.196 ↓ & 22.481±0.1 → \\
            &       & 1     & 9.3495±0.0547 & 9.2758±0.0682 ↓ & 9.3877±0.0798 ↑ & 9.2846±0.0576 ↓ & 9.314±0.0526 ↓ \\
            &       & 2     & \textbf{21.202±0.0774} & 20.98±0.118 ↓ & 21.084±0.0726 ↓ & 21.022±0.12 ↓ & 21.149±0.0641 ↓ \\
            \hhline{~-------}
            & \multirow{3}[0]{*}{80} & 0     & 22.82±0.0651 & 22.775±0.0656 ↓ & 22.803±0.0737 → & 22.832±0.0643 → & 22.811±0.0784 → \\
            &       & 1     & 12.706±0.106 & 12.498±0.157 ↓ & 12.712±0.106 → & 12.577±0.107 ↓ & 12.632±0.111 ↓ \\
            &       & 2     & 31.626±0.179 & 30.954±0.786 ↓ & 31.103±0.612 ↓ & 31.483±0.349 ↓ & 31.978±0.0964 ↑ \\
            \hhline{~-------}
            & \multirow{3}[0]{*}{100} & 0     & 16.671±0.0875 & 16.608±0.0975 ↓ & 16.693±0.061 → & 16.642±0.0952 → & 16.615±0.0886 ↓ \\
            &       & 1     & \textbf{31.291±0.106} & 31.179±0.109 ↓ & 31.25±0.111 → & 31.254±0.0838 → & 31.238±0.0947 ↓ \\
            &       & 2     & \textbf{16.025±0.241} & 15.508±0.345 ↓ & 15.955±0.293 → & 15.723±0.359 ↓ & 15.824±0.278 ↓ \\
            \hhline{~-------}
            & \multicolumn{2}{c}{\# W-D-L (VS. MPI)} &       & 0-0-12 & 1-6-5 & 0-3-9 & 1-4-7 \\
            \hhline{--------}
            \multirow{13}[0]{*}{CCP} & \multirow{3}[0]{*}{40} & 0     & \textbf{-35.019±0.124} & -35.351±0.264 ↓ & -35.198±0.169 ↓ & -35.297±0.258 ↓ & -35.141±0.164 ↓ \\
            &       & 1     & \textbf{-34.719±0.186} & -35.105±0.25 ↓ & -34.919±0.211 ↓ & -34.957±0.254 ↓ & -34.734±0.172 → \\
            &       & 2     & \textbf{-34.757±0.161} & -35.371±0.243 ↓ & -35.07±0.254 ↓ & -35.147±0.218 ↓ & -34.878±0.186 ↓ \\
            \hhline{~-------}
            & \multirow{3}[0]{*}{60} & 0     & \textbf{-53.07±0.244} & -53.895±0.459 ↓ & -53.661±0.29 ↓ & -53.589±0.287 ↓ & -53.298±0.279 ↓ \\
            &       & 1     & \textbf{-53.166±0.252} & -53.801±0.427 ↓ & -53.574±0.302 ↓ & -53.577±0.276 ↓ & -53.531±0.258 ↓ \\
            &       & 2     & \textbf{-53.105±0.347} & -53.688±0.353 ↓ & -53.547±0.362 ↓ & -53.709±0.346 ↓ & -53.309±0.281 ↓ \\
            \hhline{~-------}
            & \multirow{3}[0]{*}{80} & 0     & \textbf{-71.364±0.32} & -72.175±0.395 ↓ & -71.891±0.416 ↓ & -71.971±0.454 ↓ & -71.617±0.302 ↓ \\
            &       & 1     & -71.298±0.43 & -72.151±0.564 ↓ & -71.91±0.45 ↓ & -71.666±0.44 ↓ & \textbf{-71.287±0.432} → \\
            &       & 2     & \textbf{-70.895±0.409} & -71.703±0.611 ↓ & -71.359±0.444 ↓ & -71.53±0.41 ↓ & -71.091±0.389 ↓ \\
            \hhline{~-------}
            & \multirow{3}[0]{*}{100} & 0     & \textbf{-89.201±0.388} & -90.354±0.537 ↓ & -90.281±0.546 ↓ & -90.042±0.396 ↓ & -89.527±0.432 ↓ \\
            &       & 1     & \textbf{-88.912±0.422} & -90.41±0.575 ↓ & -89.837±0.419 ↓ & -89.737±0.478 ↓ & -89.204±0.454 ↓ \\
            &       & 2     & \textbf{-89.818±0.315} & -90.844±0.539 ↓ & -90.709±0.531 ↓ & -90.576±0.511 ↓ & -89.936±0.447 → \\
            \hhline{~-------}
            & \multicolumn{2}{c}{\# W-D-L (VS. MPI)} &       & 0-0-12 & 0-0-12 & 0-0-12 & 0-3-9 \\
            \hhline{--------}
            \multirow{13}[0]{*}{CIM} & \multirow{3}[0]{*}{40} & 0     & \textbf{35.047±0.862} & 32.309±1.79 ↓ & 34.274±0.944 ↓ & 32.62±1.52 ↓ & 34.564±0.588 ↓ \\
            &       & 1     & 22.604±3.6 & 21.884±3.79 ↓ & \textbf{26.722±1.4} ↑ & 20.114±2.74 ↓ & 21.917±3.52 ↓ \\
            &       & 2     & 23.376±1.32 & 21.703±1.3 ↓ & \textbf{23.74±0.796} → & 22.766±1.44 ↓ & 23.313±1.02 → \\
            \hhline{~-------}
            & \multirow{3}[0]{*}{60} & 0     & 49.259±1.09 & 48.561±1.18 ↓ & \textbf{49.378±0.768} → & 48.873±1.04 → & 49.178±0.8 → \\
            &       & 1     & \textbf{64.241±0.589} & 62.934±0.786 ↓ & 63.654±0.534 ↓ & 63.257±0.748 ↓ & 63.635±0.528 ↓ \\
            &       & 2     & 23.171±0.852 & 21.3±1.12 ↓ & \textbf{23.685±1.17} ↑ & 22.587±0.565 ↓ & 22.825±0.908 ↓ \\
            \hhline{~-------}
            & \multirow{3}[0]{*}{80} & 0     & \textbf{53.054±0.246} & 52.58±0.394 ↓ & 52.77±0.322 ↓ & 52.513±0.262 ↓ & 52.567±0.26 ↓ \\
            &       & 1     & \textbf{99.873±1.19} & 97.126±1.42 ↓ & 98.17±1.46 ↓ & 98.415±1.6 ↓ & 99.052±1.19 ↓ \\
            &       & 2     & 65.138±1.56 & 61.624±2.48 ↓ & \textbf{65.47±1.76} → & 63.843±1.44 ↓ & 64.123±1.5 ↓ \\
            \hhline{~-------}
            & \multirow{3}[0]{*}{100} & 0     & \textbf{113.79±1.75} & 111.2±1.63 ↓ & 111.64±1.3 ↓ & 111.39±1.23 ↓ & 113.88±1.91 → \\
            &       & 1     & \textbf{70.552±0.388} & 69.322±0.922 ↓ & 70.219±0.592 ↓ & 69.49±0.715 ↓ & 70.091±0.598 ↓ \\
            &       & 2     & \textbf{79.301±0.867} & 76.001±1.52 ↓ & 78.291±1.03 ↓ & 77.178±1.12 ↓ & 78.312±1.11 ↓ \\
            \hhline{~-------}
            & \multicolumn{2}{c}{\# W-D-L (VS. MPI)} &       & 0-0-12 & 2-3-7 & 0-1-11 & 0-3-9 \\
            \hhline{--------}
            \multirow{13}[0]{*}{CAO} & \multirow{3}[0]{*}{40} & 0     & -5185.1±3.99 & -5187.5±33.2 → & -5171.2±24.9 ↑ & -5170.1±26.7 ↑ & \textbf{-5166.9±28.3} ↑ \\
            &       & 1     & \textbf{-6827.7±17.2} & -6837.9±22.8 ↓ & -6837.9±23.5 ↓ & -6843.2±22.7 ↓ & -6829.1±20.4 → \\
            &       & 2     & -9056±0 & -9053.9±3.54 ↑ & -9054.7±2.98 → & \textbf{-9053.9±3.54} ↑ & \textbf{-9053.9±3.54} ↑ \\
            \hhline{~-------}
            & \multirow{3}[0]{*}{60} & 0     & \textbf{-8912±0} & -8931.5±20.7 ↓ & -8915.7±8.45 → & -8928.3±25.3 ↓ & -8933.9±16.1 ↓ \\
            &       & 1     & -6072±0 & -6073.3±11.3 → & -6072±0 → & -6075.7±7.65 → & \textbf{-6069.3±7.25} → \\
            &       & 2     & \textbf{-4216.5±5.44} & -4235.7±29 ↓ & -4219.5±15.8 → & -4230.4±20.4 ↓ & -4220.8±16.6 → \\
            \hhline{~-------}
            & \multirow{3}[0]{*}{80} & 0     & \textbf{-5713.1±3.99} & -5733.1±32.1 ↓ & -5714.4±4.69 → & -5717.1±7.3 ↓ & -5716±6.77 ↓ \\
            &       & 1     & \textbf{-4469.1±9.35} & -4487.7±23.2 ↓ & -4471.5±8.99 → & -4482.4±14.2 ↓ & -4469.6±7.49 → \\
            &       & 2     & -4504.3±16.3 & -4549.9±101 ↓ & -4498.7±11 → & -4515.7±21 ↓ & \textbf{-4500.3±19.5} → \\
            \hhline{~-------}
            & \multirow{3}[0]{*}{100} & 0     & \textbf{-5606.1±20.8} & -5646.4±23.5 ↓ & -5622.4±20.3 ↓ & -5623.2±24.3 ↓ & -5617.1±17.6 ↓ \\
            &       & 1     & \textbf{-4372.5±9.39} & -4409.6±68.9 ↓ & -4372.8±10.9 → & -4386.1±18.7 ↓ & -4378.7±18 → \\
            &       & 2     & \textbf{-5522.1±17.5} & -5566.4±34.4 ↓ & -5540±28.5 ↓ & -5543.2±35 ↓ & -5525.9±21.3 → \\
            \hhline{~-------}
            & \multicolumn{2}{c}{\# W-D-L (VS. MPI)} &       & 1-2-9 & 1-8-3 & 2-1-9 & 2-7-3 \\
      \hhline{--------}
      \multicolumn{3}{c}{\# Total W-D-L (VS. MPI)} &       & 1-2-69 & 6-22-44 & 2-5-65 & 7-22-43 \\
      \hhline{--------}
      \multicolumn{3}{c}{\# Highest Avg} & 50    & 0     & 13    & 2     & 8 \\
      \bottomrule[1pt]
      \end{tabular}}
    \label{tab: ga-elite detail}%
\end{table*}%

\begin{table*}[htbp]
    \centering
    \caption{Mean Value and Std of MPI and Baselines on Each Instance over 30 Runs with BRKGA. The Highest Average Performances are Highlighted in \textbf{black}. The Wilcoxon Rank-Sum Tests (\(p=0.05\)) is also Indicated by \(\uparrow\) \(\rightarrow\) and \(\downarrow\); \(\uparrow\): the Baseline Significantly Better, \(\rightarrow\): No Significant Difference, \(\uparrow\): MPI Significantly Better. \#W-D-L: Counts of \(\uparrow\) \(\rightarrow\) and \(\downarrow\). \# Highest Avg: the Number of Instances where the Method Has the Highest Average Performance.}
    \resizebox{0.75\textwidth}{!}{\begin{tabular}{cccccccc}
      \toprule[1pt]
      \makecell{Problem\\Class} & Dim & idx & MPI & Rand & SVM-SS & OBL & KAES\\
      \hhline{--------}
      \multirow{13}[0]{*}{MM} & \multirow{3}[0]{*}{40} & 0     & 39.633±0.547 & 39.5±0.619 → & 39.667±0.471 → & 39.533±0.67 → & 39.667±0.471 → \\
            &       & 1     & \textbf{39.833±0.373} & 39.467±0.562 ↓ & 39.467±0.67 ↓ & 39.433±0.667 ↓ & 39.667±0.471 → \\
            &       & 2     & 39.633±0.547 & 39.5±0.619 → & 39.367±0.706 → & 39.633±0.657 → & 39.667±0.471 → \\
            \hhline{~-------}
            & \multirow{3}[0]{*}{60} & 0     & \textbf{57.8±1.11} & 57.167±1.39 → & 57.267±1.29 → & 57.3±1.57 → & 57±1.41 ↓ \\
            &       & 1     & 57.533±1.06 & 57.2±1.35 → & 57.333±1.49 → & 57.8±1.22 → & \textbf{58.667±0.943} ↑ \\
            &       & 2     & \textbf{58.2±1.19} & 56.9±1.7 ↓ & 56.9±1.35 ↓ & 57.367±1.22 ↓ & 57.933±1.48 → \\
            \hhline{~-------}
            & \multirow{3}[0]{*}{80} & 0     & 74.667±1.3 & 73.933±2.46 → & 75.333±1.72 → & \textbf{74.833±1.86} → & 73.6±1.99 ↓ \\
            &       & 1     & \textbf{75.133±1.82} & 73.8±2.12 ↓ & 74.567±2.14 → & 73.767±2.03 ↓ & 73.233±1.31 ↓ \\
            &       & 2     & 74.4±1.52 & 73.633±2.04 → & 74.233±1.75 → & 73.267±1.46 ↓ & \textbf{75±1.63} → \\
            \hhline{~-------}
            & \multirow{3}[0]{*}{100} & 0     & \textbf{91.467±1.69} & 89.233±2.73 ↓ & 89.333±3.14 ↓ & 88.967±2.46 ↓ & 90±2.11 ↓ \\
            &       & 1     & \textbf{90.6±2.55} & 88.633±2.63 ↓ & 90.067±2.03 → & 90.067±2.37 → & 89.567±2.04 → \\
            &       & 2     & \textbf{90.7±2.48} & 88.8±2.66 ↓ & 87.633±2.59 ↓ & 89.167±2.3 ↓ & 90.6±2.36 → \\
            \hhline{~-------}
            & \multicolumn{2}{c}{\# W-D-L (VS. MPI)} &       & 0-6-6 & 0-8-4 & 0-6-6 & 1-7-4 \\
            \hhline{--------}
            \multirow{13}[0]{*}{MC} & \multirow{3}[0]{*}{40} & 0     & \textbf{218.2±3.24} & 213.57±4.65 ↓ & 216.2±4.49 ↓ & 213.6±4.21 ↓ & 212.17±2.11 ↓ \\
            &       & 1     & 303.87±4.95 & \textbf{304.97±4.1} → & 306.4±2.86 ↑ & 303.73±3.18 → & 304.27±2.52 → \\
            &       & 2     & 151.8±3.46 & 152.93±3.63 → & 156.6±3.06 ↑ & 153.7±3.25 ↑ & 153.87±2.38 ↑ \\
            \hhline{~-------}
            & \multirow{3}[0]{*}{60} & 0     & 309.4±2.58 & 309.4±4.77 → & 310.43±2.28 → & 308.7±2.75 → & 316.23±4 ↑ \\
            &       & 1     & 585.47±5.45 & 585±6.15 → & 586.8±6.79 → & 582.33±3.47 ↓ & 581.67±3.93 ↓ \\
            &       & 2     & 465.3±2.64 & 465.43±3.55 → & 473.17±7.31 ↑ & 467.23±3.49 ↑ & 466.87±6.03 → \\
            \hhline{~-------}
            & \multirow{3}[0]{*}{80} & 0     & 511.4±9.97 & 516.43±11.6 ↑ & 526.77±7.9 ↑ & 521.23±7.96 ↑ & 521.17±10.7 ↑ \\
            &       & 1     & \textbf{572.73±6.29} & 568.3±6.02 ↓ & 569.97±4.18 ↓ & 571±6.13 → & 570.83±6.09 → \\
            &       & 2     & 645.3±4.87 & 646.2±9.54 → & 648.63±4.09 ↑ & 646±8.96 → & 646.67±4.37 → \\
            \hhline{~-------}
            & \multirow{3}[0]{*}{100} & 0     & 1233.3±5.71 & 1232.9±12.8 → & 1238.9±10 ↑ & 1234.2±12.7 → & 1227.7±11.2 ↓ \\
            &       & 1     & \textbf{1342.3±8.56} & 1335.1±10.3 ↓ & 1339.4±9.89 → & 1338.2±11 → & 1333.9±10.9 ↓ \\
            &       & 2     & \textbf{1400.7±5.82} & 1393.7±8.24 ↓ & 1399.5±6.8 → & 1392±10.1 ↓ & 1395.7±11.4 ↓ \\
            \hhline{~-------}
            & \multicolumn{2}{c}{\# W-D-L (VS. MPI)} &       & 1-7-4 & 6-4-2 & 3-6-3 & 3-4-5 \\
            \hhline{--------}
            \multirow{13}[0]{*}{KP} & \multirow{3}[0]{*}{40} & 0     & \textbf{11.851±0.072} & 11.794±0.0784 ↓ & 11.779±0.104 ↓ & 11.805±0.0689 ↓ & 11.769±0.0777 ↓ \\
            &       & 1     & 12.072±0.0854 & 12.042±0.117 → & 12.014±0.0823 ↓ & 12.055±0.147 → & 12.089±0.105 → \\
            &       & 2     & \textbf{11.16±0.0528} & 11.121±0.079 ↓ & 11.119±0.0705 ↓ & 11.154±0.0922 → & 11.137±0.0904 → \\
            \hhline{~-------}
            & \multirow{3}[0]{*}{60} & 0     & 22.601±0.114 & 22.606±0.124 → & 22.599±0.0971 → & 22.623±0.131 → & 22.611±0.0873 → \\
            &       & 1     & 9.4068±0.0879 & 9.339±0.0836 ↓ & 9.4535±0.0592 → & 9.3649±0.0803 ↓ & 9.3995±0.0952 → \\
            &       & 2     & \textbf{21.289±0.0556} & 21.236±0.0815 ↓ & 21.209±0.0483 ↓ & 21.213±0.0864 ↓ & 21.198±0.0633 ↓ \\
            \hhline{~-------}
            & \multirow{3}[0]{*}{80} & 0     & 22.959±0.0783 & 22.969±0.0714 → & 22.947±0.0804 → & 22.975±0.0845 → & 22.961±0.0815 → \\
            &       & 1     & 12.795±0.0912 & 12.806±0.13 → & 12.869±0.0936 ↑ & 12.746±0.107 ↓ & 12.843±0.0991 ↑ \\
            &       & 2     & 31.959±0.117 & 31.883±0.172 → & 31.826±0.112 ↓ & 31.868±0.152 ↓ & 32.265±0.105 ↑ \\
            \hhline{~-------}
            & \multirow{3}[0]{*}{100} & 0     & \textbf{16.872±0.106} & 16.81±0.116 ↓ & 16.857±0.0704 → & 16.841±0.0878 → & 16.763±0.124 ↓ \\
            &       & 1     & \textbf{31.434±0.0655} & 31.415±0.0812 → & 31.366±0.0838 ↓ & 31.421±0.0916 → & 31.425±0.0704 → \\
            &       & 2     & 16.542±0.263 & 16.348±0.369 ↓ & 16.603±0.34 → & 16.462±0.297 → & 16.41±0.25 ↓ \\
            \hhline{~-------}
            & \multicolumn{2}{c}{\# W-D-L (VS. MPI)} &       & 0-6-6 & 1-5-6 & 0-7-5 & 2-6-4 \\
            \hhline{--------}
            \multirow{13}[0]{*}{CCP} & \multirow{3}[0]{*}{40} & 0     & \textbf{-34.886±0.112} & -34.967±0.153 ↓ & -34.995±0.184 ↓ & -34.964±0.18 ↓ & -34.897±0.154 → \\
            &       & 1     & \textbf{-34.443±0.125} & -34.606±0.0968 ↓ & -34.54±0.14 ↓ & -34.495±0.141 → & -34.494±0.113 → \\
            &       & 2     & \textbf{-34.504±0.113} & -34.664±0.19 ↓ & -34.613±0.172 ↓ & -34.663±0.162 ↓ & -34.6±0.128 ↓ \\
            \hhline{~-------}
            & \multirow{3}[0]{*}{60} & 0     & \textbf{-52.63±0.219} & -52.696±0.221 → & -52.692±0.209 → & -52.759±0.193 ↓ & -52.631±0.253 → \\
            &       & 1     & -52.773±0.174 & \textbf{-52.723±0.208} → & -52.738±0.195 → & -52.741±0.194 → & -52.861±0.179 ↓ \\
            &       & 2     & \textbf{-52.51±0.169} & -52.688±0.207 ↓ & -52.638±0.215 ↓ & -52.756±0.233 ↓ & -52.585±0.217 → \\
            \hhline{~-------}
            & \multirow{3}[0]{*}{80} & 0     & \textbf{-70.481±0.196} & -70.643±0.215 ↓ & -70.68±0.375 ↓ & -70.642±0.316 ↓ & -70.643±0.233 ↓ \\
            &       & 1     & -70.452±0.204 & -70.51±0.207 → & -70.724±0.199 ↓ & -70.61±0.264 ↓ & \textbf{-70.226±0.256} ↑ \\
            &       & 2     & -70.182±0.265 & -70.202±0.279 → & -70.145±0.341 → & \textbf{-70.094±0.271} → & -70.115±0.26 → \\
            \hhline{~-------}
            & \multirow{3}[0]{*}{100} & 0     & -88.059±0.316 & -88.105±0.448 → & -88.144±0.396 → & \textbf{-87.873±0.358} ↑ & -87.985±0.269 → \\
            &       & 1     & -87.778±0.274 & -88.178±0.39 ↓ & -88.236±0.406 ↓ & -88.003±0.335 ↓ & \textbf{-87.676±0.279} → \\
            &       & 2     & \textbf{-88.653±0.288} & -88.856±0.276 ↓ & -88.952±0.324 ↓ & -88.896±0.281 ↓ & -88.694±0.339 → \\
            \hhline{~-------}
            & \multicolumn{2}{c}{\# W-D-L (VS. MPI)} &       & 0-5-7 & 0-4-8 & 1-3-8 & 1-8-3 \\
            \hhline{--------}
            \multirow{13}[0]{*}{CIM} & \multirow{3}[0]{*}{40} & 0     & \textbf{35.847±0.696} & 34.91±0.952 ↓ & 35.478±0.855 ↓ & 34.828±0.722 ↓ & 35.555±0.599 ↓ \\
            &       & 1     & 24.862±4.15 & 23.437±4.41 ↓ & 28.602±0.714 ↑ & 22.056±4.2 ↓ & 25.857±4.01 → \\
            &       & 2     & 24.116±1.72 & 24.504±1.56 → & 25.041±0.887 → & 24.28±1.68 → & 24.959±1.01 → \\
            \hhline{~-------}
            & \multirow{3}[0]{*}{60} & 0     & 51.434±0.863 & 51.151±0.545 ↓ & 51.474±0.783 → & 51.171±0.813 → & 51.124±0.683 ↓ \\
            &       & 1     & \textbf{65.239±0.648} & 65.191±0.545 → & 65.138±0.561 → & 65.1±0.477 → & 65.128±0.51 → \\
            &       & 2     & 23.97±0.69 & 23.929±0.881 → & 27.359±3.83 ↑ & 24.471±2.26 → & 24.392±1.81 → \\
            \hhline{~-------}
            & \multirow{3}[0]{*}{80} & 0     & \textbf{53.376±0.172} & 53.297±0.19 → & 53.184±0.181 ↓ & 53.264±0.195 ↓ & 53.225±0.158 ↓ \\
            &       & 1     & \textbf{102.95±0.685} & 102.62±0.86 ↓ & 102.79±0.977 → & 102.76±0.857 → & 102.7±0.815 → \\
            &       & 2     & 66.346±1.78 & 67.065±1.83 ↑ & 67.798±1.99 ↑ & 66.55±2.47 → & 67.446±1.87 ↑ \\
            \hhline{~-------}
            & \multirow{3}[0]{*}{100} & 0     & 117.77±1.11 & 117.18±1.38 ↓ & 116.84±1.16 ↓ & 117.07±1.06 ↓ & 117.12±1.46 → \\
            &       & 1     & \textbf{71.249±0.309} & 71.08±0.359 ↓ & 71.051±0.395 ↓ & 71.293±0.279 → & 71.122±0.287 → \\
            &       & 2     & \textbf{82.133±0.776} & 81.377±1.01 ↓ & 81.63±0.929 ↓ & 81.278±0.87 ↓ & 81.863±0.791 → \\
            \hhline{~-------}
            & \multicolumn{2}{c}{\# W-D-L (VS. MPI)} &       & 1-4-7 & 3-4-5 & 0-7-5 & 1-8-3 \\
            \hhline{--------}
            \multirow{13}[0]{*}{CAO} & \multirow{3}[0]{*}{40} & 0     & -5184±0 & \textbf{-5162.7±30.2} ↑ & -5168±26.8 ↑ & \textbf{-5162.7±30.2} ↑ & -5166.9±28.3 ↑ \\
            &       & 1     & \textbf{-6816.5±2.87} & -6838.4±25 ↓ & -6830.7±21.9 ↓ & -6832±22.6 ↓ & -6838.7±28.2 ↓ \\
            &       & 2     & -9056±0 & -9053.6±3.67 ↑ & -9053.1±3.86 ↑ & -9053.9±3.54 ↑ & \textbf{-9052.8±3.92} ↑ \\
            \hhline{~-------}
            & \multirow{3}[0]{*}{60} & 0     & \textbf{-8912±0} & -8921.9±14.8 ↓ & -8919.2±11 ↓ & -8922.4±11.9 ↓ & -8926.7±15.6 ↓ \\
            &       & 1     & -6070.9±2.72 & -6070.1±6.09 → & -6071.2±4.31 → & \textbf{-6069.9±6.17} → & -6071.7±1.44 → \\
            &       & 2     & \textbf{-4211.2±3.92} & -4221.3±19.6 → & -4217.1±16.6 → & -4212.5±13.2 → & -4215.7±18.1 → \\
            \hhline{~-------}
            & \multirow{3}[0]{*}{80} & 0     & \textbf{-5712±0} & -5714.7±4.3 ↓ & -5714.7±4.77 ↓ & -5714.4±3.67 ↓ & -5715.7±3.99 ↓ \\
            &       & 1     & \textbf{-4466.1±5.44} & -4467.7±6.77 → & -4469.9±7.71 → & -4468.3±7.08 → & -4468.8±7.33 → \\
            &       & 2     & -4491.5±5.34 & -4493.1±8.64 → & -4489.6±9.99 → & -4490.4±9.72 → & \textbf{-4482.1±11.5} ↑ \\
            \hhline{~-------}
            & \multirow{3}[0]{*}{100} & 0     & \textbf{-5601.3±17.8} & -5612.8±18.9 ↓ & -5610.1±15.6 ↓ & -5613.1±19 ↓ & -5610.1±18.8 ↓ \\
            &       & 1     & -4368.5±2.87 & -4369.9±7.64 → & \textbf{-4368±0} → & -4369.1±3.99 → & -4368.5±2.87 → \\
            &       & 2     & \textbf{-5504±0} & -5506.9±10.4 → & -5509.6±11.2 ↓ & -5509.6±14 → & -5505.6±10.2 → \\
            \hhline{~-------}
            & \multicolumn{2}{c}{\# W-D-L (VS. MPI)} &       & 2-6-4 & 2-5-5 & 2-6-4 & 3-5-4 \\
      \hhline{--------}
      \multicolumn{3}{c}{\# Total W-D-L (VS. MPI)} &       & 4-34-34 & 12-30-30 & 6-35-31 & 11-28-23 \\
      \hhline{--------}
      \multicolumn{3}{c}{\# Highest Avg} & 36    & 3     & 16    & 7     & 11 \\
      \bottomrule[1pt]
    \end{tabular}}
    \label{tab: brkga detail}%
\end{table*}%

\section{Detailed Experimental Results on Each ProblemInstance}
The detailed experimental results on each problem instance are presented in Table~\ref{tab: ga-elite detail} and Table~\ref{tab: brkga detail}. Table~\ref{tab: ga-elite detail} presents the results with GA-Elite and Table~\ref{tab: brkga detail} presents the result with BRKGA.




\end{document}